\documentclass[final]{cvpr}
\usepackage{times}
\usepackage{epsfig}
\usepackage{graphicx}
\usepackage{amsmath}
\usepackage{amssymb}
\usepackage{listings}
\usepackage{enumerate}
\usepackage{pbox}
\usepackage{url}
\usepackage{mathtools}
\usepackage{amsmath}
\usepackage{amssymb}

\usepackage{blindtext}
\usepackage{algorithm}
\usepackage{algpseudocode}

\usepackage{epsfig}
\usepackage{graphicx}
\usepackage{float}
\usepackage{subcaption}
\usepackage{caption}
\usepackage{makecell}

\usepackage{epsfig}
\usepackage{graphicx}
\usepackage{subcaption}
\usepackage{enumerate}

\usepackage{booktabs}
\usepackage{pifont}
\newcommand{\cmark}{\ding{51}}%
\newcommand{\xmark}{\ding{55}}%

\usepackage[table,x11names]{xcolor}
\usepackage{multicol}
\usepackage{multirow}

\usepackage[misc]{ifsym}

\newcommand{\comment}[1]{}

\renewcommand{\Sigma}{\mathfrak{S}}

\newlength\savewidth\newcommand\shline{\noalign{\global\savewidth\arrayrulewidth
  \global\arrayrulewidth 1pt}\hline\noalign{\global\arrayrulewidth\savewidth}}
\newcommand\paperurl[1]{{\footnotesize{\color{blue}{\url{#1}}}}}

\makeatletter
\def\@fnsymbol#1{\ensuremath{\ifcase#1\or \dagger\or \dagger\or
\mathsection\or \mathparagraph\or \|\or **\or \dagger\dagger
\or \ddagger\ddagger \else\@ctrerr\fi}}
\makeatother

\newcommand{\tablestyle}[2]{\ttfamily\setlength{\tabcolsep}{#1}\renewcommand{\arraystretch}{#2}\centering\footnotesize}

\usepackage[pagebackref=true,breaklinks=true,letterpaper=true,colorlinks,bookmarks=false]{hyperref}

\usepackage{algorithm}
\usepackage[]{algpseudocode}

\begin{document}

\title{DETR Doesn't Need Multi-Scale or Locality Design}

\author{
Yutong Lin$^{1}{\thanks{Equal contribution. \Letter\; \texttt{\{yuhui.yuan, hanhu\}@microsoft.com}}}$\quad\:
Yuhui Yuan$^{2\dagger}$ \quad
Zheng Zhang$^{2\dagger}$ \quad
Chen Li$^1$ \quad
Nanning Zheng$^1$ \quad
Han Hu$^{2\dagger}$ \\[2mm]
$^1$Xi'an Jiaotong University \quad\quad
$^2$Microsoft Research Asia
}

\maketitle

\begin{abstract}
This paper presents an improved DETR detector that maintains a ``plain'' nature: using a single-scale feature map and global cross-attention calculations without specific locality constraints, in contrast to previous leading DETR-based detectors that reintroduce architectural inductive biases of multi-scale and locality into the decoder. We show that two simple technologies are surprisingly effective within a plain design to compensate for the lack of multi-scale feature maps and locality constraints. The first is a box-to-pixel relative position bias (BoxRPB) term added to the cross-attention formulation, which well guides each query to attend to the corresponding object region while also providing encoding flexibility. The second is masked image modeling (MIM)-based backbone pre-training which helps learn representation with fine-grained localization ability and proves crucial for remedying dependencies on the multi-scale feature maps.
By incorporating these technologies and recent advancements in training and problem formation, the improved ``plain'' DETR showed exceptional improvements over the original DETR detector. By leveraging the Object365 dataset for pre-training, it achieved 63.9 mAP accuracy using a Swin-L backbone, which is highly competitive with state-of-the-art detectors which all heavily rely on multi-scale feature maps and region-based feature extraction. Code will be available at 
{\url{{https://github.com/impiga/Plain-DETR}}}.
\end{abstract}

\section{Introduction}

The recent revolutionary advancements in natural language processing highlight the importance of keeping task-specific heads or decoders as general, simple, and lightweight as possible, and shifting main efforts towards building more powerful large-scale foundation models~\cite{radford2018improving,devlin2018bert,brown2020language}. However, the computer vision community often continues to focus heavily on the tuning and complexity of task-specific heads, resulting in designs that are increasingly heavy and complex. 

The development of DETR-based object detection methods follows this trajectory. The original DETR approach~\cite{carion2020end} is impressive in that it discarded complex and domain-specific designs such as multi-scale feature maps and region-based feature extraction that require a dedicated understanding of the specific object detection problem. Yet, subsequent developments~\cite{zhu2020deformable,zhang2022dino} in the field have reintroduced these designs, which do improve training speed and accuracy but also contravene the principle of ``fewer inductive biases''~\cite{vit20}.

In this work, we aim to improve upon the original DETR detector, while preserving its ``plain'' nature: \emph{no multi-scale feature maps}, \emph{no locality design for cross-attention calculation}. This is challenging as object detectors need to handle objects of varying scales and locations. Despite the latest improvements in training and problem formulation, as shown in Table~\ref{tab:exp_modern_detr}, the plain DETR method still lags greatly behind state-of-the-art detectors that utilize multi-scale feature maps and regional feature extraction design.

\begin{figure}[t]
\centering
\includegraphics[width=0.455\textwidth]{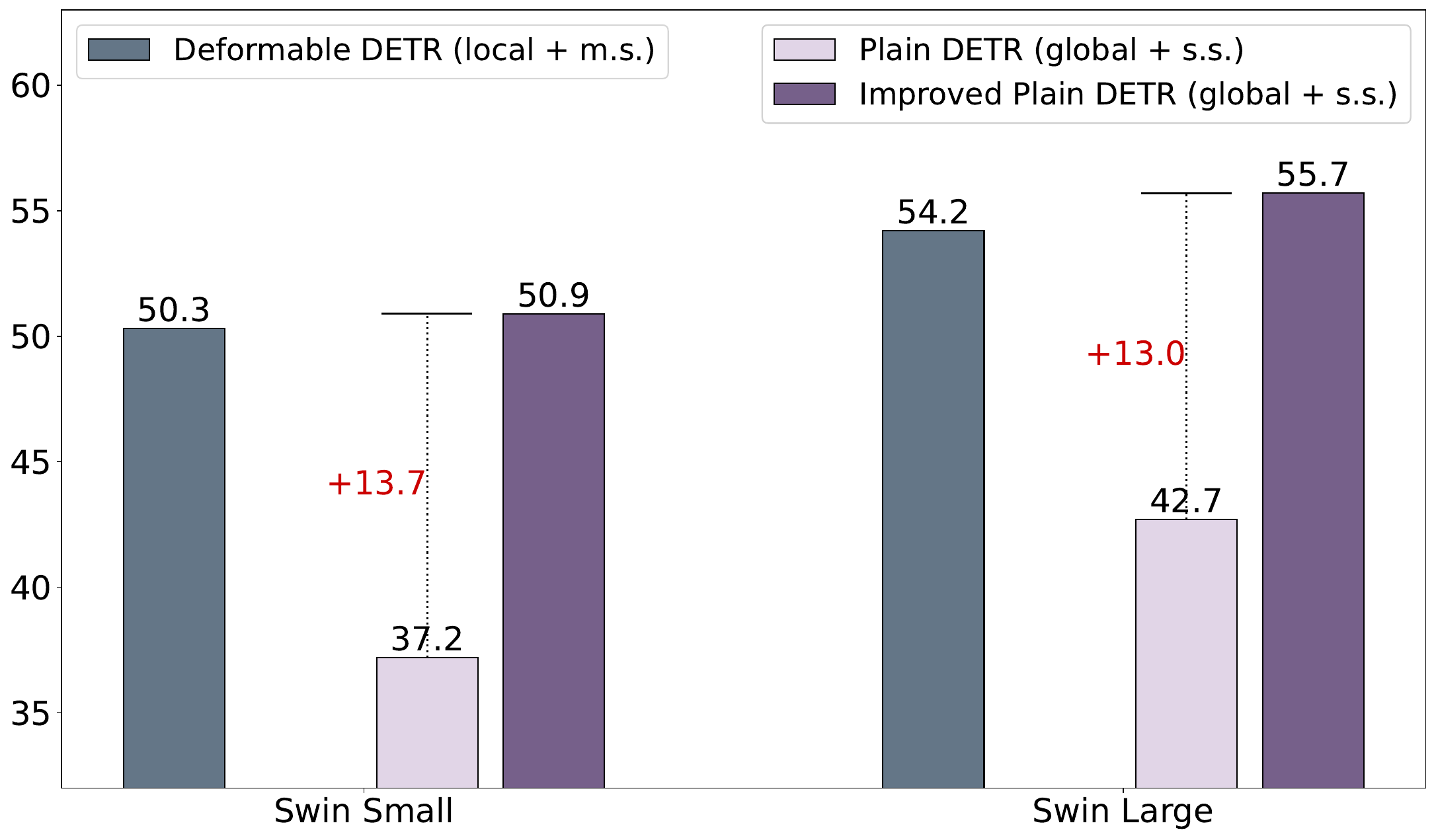}
\caption{We improve the plain DETR detectors, which rely on global cross-attention calculation and single-scale (s.s.) feature maps, by huge margins, using both Swin-S and Swin-L backbones. It makes plain DETRs as competitive as the present leading DETR detectors based on local cross-attention and multi-scale (m.s.) feature maps.}
\label{fig:intro}
\vspace{-5mm}
\end{figure}

So, how can we compensate for these architectural ``inductive biases'' in addressing multi-scale and arbitrarily located objects? Our exploration found that two simple technologies, though not entirely new, were surprisingly effective in this context: box-to-pixel relative position bias (BoxRPB) and masked image modeling (MIM) pre-training. BoxRPB is inspired by the relative position bias (RPB) term in vision Transformers~\cite{liu2021swin,liu2022swin} which encodes the geometric relationship between pixels and enhances translation invariance. BoxRPB extends RPB to encode the geometric relationship between 4$d$- boxes and 2$d$- pixels. We also present an axial decomposition approach for efficient computation, with no loss of accuracy compared to using the full term. Our experiments show that the BoxRPB term can well guide the cross-attention computation to be well dedicated to individual objects (see Figure~\ref{fig:cross_attn}, and it dramatically improves detection accuracy by +8.9 mAP over a plain DETR baseline of 37.2 mAP on the COCO benchmark (see Table~\ref{tab:approach_ablation}). 

The utilization of MIM pre-training is another crucial technology in enhancing the performance of plain DETR. Our results demonstrate also a significant improvement of +7.4 mAP over the plain DETR baseline (see Table~\ref{tab:approach_ablation}), which may be attributed to its fine-grained localization capability~\cite{xie2022revealing}. While MIM pre-training has been shown to moderately improve the performance of other detectors~\cite{he2022masked,xie2021simmim}, its impact in plain settings is profound. Furthermore, the technology has proven to be a key factor in eliminating the necessity of using multi-scale feature maps from the backbones, thereby expanding the findings in~\cite{li2022exploring,MIMDet} to detectors that utilize hierarchical backbones or single-scale heads.

By incorporating these technologies and the latest improvements in both training and problem formulation, our improved ``plain'' DETR has demonstrated exceptional improvements over the original DETR detector, as illustrated in Figure~\ref{fig:intro}. Furthermore, our method achieved an accuracy of 63.9 mAP when utilizing the Object365 dataset for pre-training, making it highly competitive with state-of-the-art object detectors that rely on multi-scale feature maps and region-based feature extraction techniques, such as cascade R-CNN~\cite{liu2022swin} and DINO~\cite{zhang2022dino}, among others.

Beyond these outcomes, our methodology exemplifies how to minimize the architectural ``inductive bias'' when designing an effective task-specific head or decoder, as opposed to relying on detection-specific multi-scale and localized designs. Our study hopes to inspire future research on using generic plain decoders, such as that of DETR, for a wider range of visual problems with minimal effort, thus allowing the field to shift more energy to developing large foundation visual models, similar to what occurs in the field of natural language processing.

\section{A Modernized Plain DETR Baseline}

\subsection{A Review of the Original DETR}

\vspace{1mm}

The original DETR detector~\cite{carion2020end} is consist of 3 sub-networks: 
\begin{itemize}
\item \emph{A backbone network} $\mathcal{F}_b$ to extract image features from an image. We denote the input image as $\mathbf{I} {\in} \mathbb{R}^{\mathsf{H}\times \mathsf{W}\times \mathsf{3}}$. The backbone network can provide multi-scale feature maps ${\mathbf{C}^{2},\mathbf{C}^{3},\mathbf{C}^{4}, \mathbf{C}^{5}}$, if a convectional ConvNet is used, i.e., ResNet~\cite{He2016ResNet}. The spatial resolutions are typically ${1}/{4^2}$, ${1}/{8^2}$, ${1}/{16^2}$, and ${1}/{32^2}$ of the input image. The original DETR detector used the mainstream backbone architecture at the time, ResNet, as its backbone network, and either an original ResNet or a variant with a dilated stage 5 network is used.
Now the mainstream backbone network has evolved to vision Transformers, which will be used in our experiments, e.g., Swin Transformer~\cite{liu2021swin}.

\item \emph{A Transformer encoder} $\mathcal{F}_e$ to enhance the image features. It applies on $\mathbf{P}^{{5}}\in \mathbb{R}^{\frac{\mathsf{HW}}{32^2}\times \mathsf{C}}$ ($\mathsf{C}$=$256$), obtained via a linear projection on $\mathbf{C}^{5}$. The Transformer encoder usually consists of several stacking Transformer blocks, i.e., 6 in the original DETR.
\item \emph{A global Transformer decoder} $\mathcal{F}_d$ to decode object bounding boxes from the image feature map using a set of randomly initialized object queries $\mathbf{Q}=\{\mathbf{q}_0,\mathbf{q}_1,\cdots,\mathbf{q}_n\}$. The Transformer decoder also usually consists of multiple layers, with each layer including a self-attention block, a cross-attention block, and a feed-forward block. Each of the decoder layers will produce a set of objects with labels and bounding boxes, driven by a set matching loss.
\end{itemize}
 
The DETR framework possesses several merits, including: 1) Conceptually straightforward and generic in applicability. It views object detection as a pixel-to-object ``translation'' task, with a generic notion of decoding image pixels into problem targets. 2) Requiring minimal domain knowledge, such as custom label assignments and hand-designed non-maximum suppression, due to the use of an end-to-end set matching loss. 3) Being plain, avoiding domain-specific multi-scale feature maps and region-based feature extraction.

In the following, we will first build an enhanced DETR-based detector by incorporating recent advancements regarding both training and problem formulation, while maintaining the above nice merits.

\subsection{An Enhanced Plain DETR Baseline}\label{sec:strong_baseline}

\vspace{1mm}
\noindent \textbf{Basic setup.} 
Our basic setup mostly follows the original DETR framework, except for the following adaptations: 1) We use a stronger Swin-T backbone, instead of the original ResNet50 backbone; 2) We create a feature map of $\mathbf{P}_4$ from $\mathbf{C}_5$ by deconvolution, instead of adding dilation operations to the last stage of the backbone, for simplicity purpose. 3) We set the number of queries as 300, and the dropout ratio of the Transformer decoder as 0. 4) We use $1\times$ scheduler settings (12 epochs) for efficient ablation study. As shown in Table~\ref{tab:exp_modern_detr}, this basic setup produces a 22.5 mAP on COCO \texttt{val}. 

In the following, we incorporate some recent advancements in training and problem formulation into the basic setup, and gradually improve the detection accuracy to 37.2 mAP, as shown in Table~\ref{tab:exp_modern_detr}.

\vspace{1mm}
\noindent \textbf{Merging Transformer encoder into the backbone.} The backbone network and Transformer encoder serve the same purpose of encoding image features. We discovered that by utilizing a Vision Transformer backbone, we are able to consolidate the computation budget of the Transformer encoder into the backbone, with slight improvement, probably because more parameters are pre-trained. Specifically, we employed a Swin-S backbone and removed the Transformer encoder. This method resulted in similar computation FLOPs compared to the original Swin-T plus 6-layer Transformer encoder. This approach simplifies the overall DETR framework to consist of only a backbone (encoder) and a decoder network.

\vspace{1mm}
\noindent \textbf{Focal loss for better classification}. We follow~\cite{zhu2020deformable} to utilize focal loss~\cite{lin2017focal} to replace the default cross-entropy loss, which improves the detection accuracy significantly from 23.1 mAP to 31.6 mAP.

\vspace{1mm}
\noindent \textbf{Iterative refinement.}
We follow the iterative refinement scheme~\cite{teed2020raft,zhu2020deformable,cai2018cascade} to make each decoder layer predict the box delta over the latest bounding box produced by the previous decoder layer, unlike the original DETR that uses independent predictions within each Transformer decoder layer. This strategy improves the detection accuracy by +1.5 mAP to reach 33.1 mAP.

\vspace{1mm}
\noindent \textbf{Content-related query.} We follow~\cite{zhu2020deformable} to generate object queries based on image content. The top 300 most confident predictions are selected as queries for the subsequent decoding process. A set matching loss is used for object query generation, thereby maintaining the merit of  no domain-specific label assignment strategy. This modification resulted in a +0.9 mAP improvement in detection accuracy, reaching 34.0 mAP.

\vspace{1mm}
\noindent \textbf{Look forward twice.}
We incorporate the look forward twice scheme~\cite{zhang2022dino,jia2022detrs} to take advantage of the refined box information from previous Transformer decoder layers, thereby more effectively optimizing the parameters across adjacent Transformer decoder layers. This modification yields +0.8 mAP improvements.

\vspace{1mm}
\noindent \textbf{Mixed query selection.} This method~\cite{zhang2022dino} combines the static content queries with image-adaptive position queries to form better query representations. It yields +0.4 mAP improvements.

\vspace{1mm}
\noindent \textbf{Hybrid matching.} The original one-to-one set matching is less efficacy in training positive samples. There have been several methods to improve the efficacy through an auxiliary one-to-many set matching loss~\cite{jia2022detrs, chen2022group, li2022dn}. We opted for the hybrid matching approach~\cite{jia2022detrs}, as it preserves the advantage of not requiring additional manual labeling noise or assignment designs. This modification resulted in a +2.0 mAP improvement in detection accuracy, achieving a final 37.2 mAP.

\begin{table}[t]
\begin{minipage}[t]{1\linewidth}
\begin{minipage}[t]{1\linewidth}
\vspace{2mm}
\centering
\setlength{\tabcolsep}{8pt}
\footnotesize
\renewcommand{\arraystretch}{1.2}
\resizebox{1.0\linewidth}{!}
{
    \begin{tabular}{c|c|c|c|c|c|c|c}
        MTE  & FL  & IR  &  TS  &  LFT & MQS & HM  & AP                 \\
        \shline
        \xmark       & \xmark & \xmark & \xmark & \xmark & \xmark & \xmark & $22.5$             \\
        \cmark       & \xmark & \xmark & \xmark & \xmark & \xmark & \xmark & $23.1$             \\
        \cmark       & \cmark & \xmark & \xmark & \xmark & \xmark & \xmark & $31.6$             \\
        \cmark       & \cmark & \cmark & \xmark & \xmark & \xmark & \xmark & $33.1$             \\
        \cmark       & \cmark & \cmark & \cmark & \xmark & \xmark & \xmark & $34.0$             \\
        \cmark       & \cmark & \cmark & \cmark & \cmark & \xmark & \xmark & $34.8$             \\
        \cmark       & \cmark & \cmark & \cmark & \cmark & \cmark & \xmark & $35.2$             \\
        \cmark       & \cmark & \cmark & \cmark & \cmark & \cmark & \cmark & $\bf{37.2}$ \\
    \end{tabular}
}
\caption{\small{\textbf{Preliminary ablation results on the effect of each factor that is used to modernize plain DETR.}} MTE: merging the Transformer encoder. FL: classification loss as a focal loss. IR: Iterative refinement. TS: two-stage. LFT: look forward twice. MQS: mixed query selection. HM: hybrid matching.}
\label{tab:exp_modern_detr}
\end{minipage}
\end{minipage}
\end{table}

\section{Box-to-Pixel Relative Position Bias}
\label{sec:our_approach}

In this section, we introduce a simple technology, box-to-pixel relative position bias (BoxRPB), that proves critical to compensate for the lack of multi-scale features and the explicit local cross-attention calculations.

The original DETR decoder adopts a standard cross-attention computation:
\begin{align}
\label{eq.general_cross_attn}
  \mathbf{O} = \operatorname{Softmax}(\mathbf{Q}\mathbf{K}^{\text{T}})\mathbf{V} + \mathbf{X},
\end{align}
where $X$ and $O$ are the input and output features of each object query, respectively; $Q$, $K$ and $V$ are query, key, and value features, respectively.

As will be shown in Figure~\ref{fig:cross_attn}, the original cross-attention formulation often attends to irrelevant image areas within a plain DETR framework. We conjecture that this may be a reason for its much lower accuracy than that with multi-scale and explicit locality designs. Inspired by the success of pixel-to-pixel relative position bias for vision Transformer architectures~\cite{liu2021swin,liu2022swin}, we explore the use of box-to-pixel relative position bias (BoxRPB) for cross-attention calculation:
\begin{align}
\label{eq.general_cross_attn}
  \mathbf{O} = \operatorname{Softmax}(\mathbf{Q}\mathbf{K}^{\text{T}}\textcolor{red}{\;+\;\mathbf{B}})\mathbf{V} + \mathbf{X},
\end{align}
where $\mathbf{B}$ is the relative position bias determined by the geometric relationship between boxes and pixels.

Different from the original relative position bias (RPB) which is defined on 2$d$- relative positions, the BoxRPB needs to handle a larger geometric space of 4$d$. In the following, we introduce two implementation variants.

\vspace{2mm}
\noindent\textbf{A Naive BoxRPB implementation.}
We adapt the continuous RPB method ~\cite{liu2022swin} to compute the 4$d$- box-to-pixel relative position bias. The original continuous RPB method~\cite{liu2022swin} produces the bias term for each relative position configuration by a meta-network applied on the corresponding 2$d$- relative coordinates. When extending this method for BoxRPB, we use the top-left and bottom-right corners to represent a box and use the relative positions between these corner points and the image pixel point as input to the meta-network. Denote the relative coordinates as $(\Delta\mathbf{x}_1, \Delta\mathbf{y}_1)\in \mathbb{R}^{\mathsf{K}\times\mathsf{H}\times\mathsf{W}\times2} $ and $(\Delta\mathbf{x}_2, \Delta\mathbf{y}_2)\in \mathbb{R}^{\mathsf{K}\times\mathsf{H}\times\mathsf{W}\times2}$, the box-to-pixel relative position bias can be defined as:
\begin{equation}
\begin{aligned}
\label{eq.rpe_mlp}
{\mathbf{B}} =\operatorname{MLP}(\Delta\mathbf{x}_1, \Delta\mathbf{y}_1, \Delta\mathbf{x}_2, \Delta\mathbf{y}_2),
\end{aligned}
\end{equation}
where $\mathbf{B}$ is in a shape of $\mathbb{R}^{\mathsf{K}\times\mathsf{W}\mathsf{H}\times\mathsf{M}}$, with $\mathsf{M}$ denoting the number of attention heads, $\mathsf{K}$ denoting the number of predicted bounding boxes, $\mathsf{W}$, $\mathsf{H}$ denoting the width and  height of the output feature maps; the MLP network consists of two linear layers: $\operatorname{Linear}\to\operatorname{ReLU}\to\operatorname{Linear}$. The input/output shapes of these two linear layers are: $\mathsf{K}{\times}\mathsf{H}{\times}\mathsf{W}{\times4}${$\to$}$\mathsf{K}{\times}\mathsf{H}{\times}\mathsf{W}{\times}256$ and $\mathsf{K}{\times}\mathsf{H}{\times}\mathsf{W}{\times}256${$\to$}$\mathsf{K}{\times}\mathsf{H}{\times}\mathsf{W}{\times}\mathsf{M}$, respectively.

Our experiments show that this naive implementation already performs very effectively, as shown in Table~\ref{tab:box_rpb_ablation:decomp}. However, it will consume a lot of GPU computation and memory budget and thus is not practical.

\begin{figure*}[t]
\centering
\includegraphics[width=\textwidth]{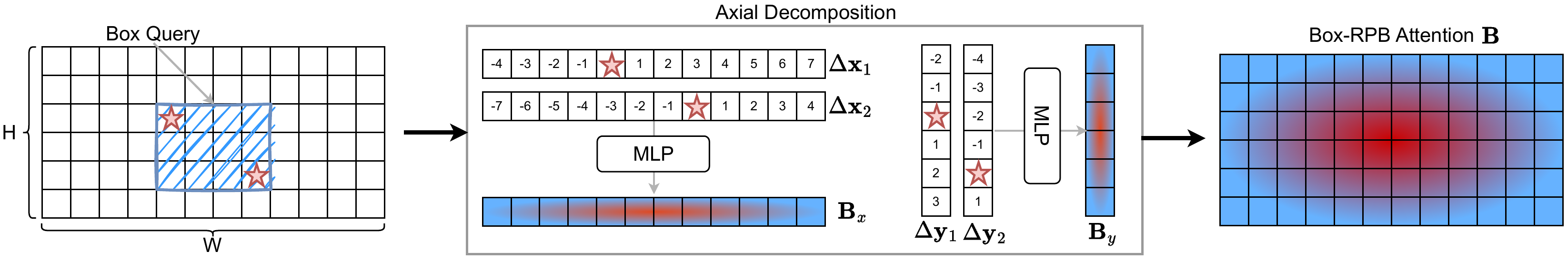}
\caption{\small{\textbf{Illustrating the details of the proposed BoxRPB scheme.}
\textit{(Left)}: The black grid represents an input image. The blue sketch region represents a predicted bounding box. We mark the top-left and right-down corners of the box with red stars. \textit{(Middle)}: Our BoxRPB calculates the offsets between all positions and the two corners along both $x$-axis and $y$-axis. Then, we concatenate the offset vectors along each axis to form ($\Delta\mathbf{x}_1$, $\Delta\mathbf{x}_2$) and ($\Delta\mathbf{y}_1$, $\Delta\mathbf{y}_2$) and apply an independent MLP to obtain the relative position bias terms ${\mathbf{B}_x}$ and ${\mathbf{B}_y}$. \textit{(Right)}: We broadcast and add ${\mathbf{B}_x}$ to ${\mathbf{B}_y}$ to get the 2D relative bias term $\mathbf{B}$. We color the positions with higher attention values with red color and blue color otherwise.
}}
\label{fig:axial_rpb}
\end{figure*}

\vspace{2mm}
\noindent\textbf{A decomposed BoxRPB implementation.} Now, we present a more efficient implementation of BoxRPB. Instead of directly computing the bias term for a 4$d$- input, we consider decomposing the bias computation into two terms:
\begin{equation}
\begin{aligned}
\label{eq.rpe_mlp_broadcast}
{\mathbf{B}} = \operatorname{{unsqueeze}}(\mathbf{B}_x, 1) + \operatorname{{unsqueeze}}(\mathbf{B}_y, 2),
\end{aligned}
\end{equation}
where $\mathbf{B}_x\in \mathbb{R}^{\mathsf{K}\times\mathsf{W}\times\mathsf{M}}$ and $\mathbf{B}_y\in \mathbb{R}^{\mathsf{K}\times\mathsf{H}\times\mathsf{M}}$ are the biases regarding $x$- axis and $y$- axis, respectively. They are computed as:
\begin{equation}
\begin{aligned}
\label{eq.rpe_mlp}
{\mathbf{B}_x} = \operatorname{MLP_1}({\Delta\mathbf{x}_1, \Delta\mathbf{x}_2}), \quad
{\mathbf{B}_y} = \operatorname{MLP_2}({\Delta\mathbf{y}_1, \Delta\mathbf{y}_2}),
\end{aligned}
\end{equation}
The overall process of the decomposed BoxRPB implementation is also illustrated in Figure~\ref{fig:axial_rpb}.
The input/output shapes of the two linear layers within $\operatorname{MLP_1}$ are: $\mathsf{K}{\times}\mathsf{W}{\times}2${$\to$}$\mathsf{K}{\times}\mathsf{W}{\times}\mathsf{256}$ and $\mathsf{K}{\times}\mathsf{W}{\times}\mathsf{256}${$\to$}$\mathsf{K}{\times}\mathsf{W}{\times}\mathsf{M}$, respectively. Similarly, the input/output shapes for the two linear layers within $\operatorname{MLP_2}$ follow the same pattern.

Through decomposition, both the computation FLOPs and memory consumption are significantly reduced, while the accuracy almost keeps, as shown in Table~\ref{tab:box_rpb_ablation:decomp}. This decomposition-based implementation is used default in our experiments.

Figure~\ref{fig:cross_attn} shows the effect of this additional BoxRPB term for cross-attention computation. In general, the BoxRPB term makes the attention focused more on the objects and box boundaries, while the cross-attention without the BoxRPB may attend to many irrelevant areas. This may explain the significantly improved accuracy (+8.9 mAP) by the BoxRPB term, as shown in Table~\ref{tab:approach_ablation}.

\section{More Improvements}
\vspace{1mm}

In this section, we introduce two other technologies that can additionally improve the plain DETR framework. 

\noindent\textbf{MIM pre-training.}
We leverage the recent advances of masked image modeling pre-training\cite{bao2021beit,he2022masked,xie2022simmim,li2022exploring} which have shown better locality\cite{xie2022revealing}. Specifically, we initialize the Swin transformer backbone with SimMIM pre-trained weights that are learned on ImageNet without labels as in\cite{xie2022simmim}. 

As shown in Table~\ref{tab:approach_ablation}, the MIM pre-trainig brings +7.4 mAP improvements over our plain DETR baseline. The profound gains of MIM pre-training on the plain DETR framework than on other detectors may highlight the importance of the learned localization ability for a plain DETR framework. On a higher baseline where BoxRPB has been involved, the MIM pre-training can still yield +2.6 mAP gains, reaching 48.7 mAP. Moreover, we note that MIM pre-training is also crucial for enabling us abandon the multi-scale backbone features with almost no loss of accuracy, as shown by Table~\ref{tab:mim_ablation:2} and \ref{tab:mim_ablation:3}.

\vspace{1mm}
\noindent\textbf{Bounding box regression with re-parameterization.} Another improvement we would like to highlight is the bounding box re-parameterization when performing bounding box regression.

The original DETR framework~\cite{carion2020end} and most of its variants directly scale the box centers and sizes to [$0$,$1$]. It will face difficulty in detecting small objects due to the large objects dominating the loss computation. Instead, we re-parameterize the box centers and sizes of $l$-th decoder layer as:
\begin{equation}
\begin{aligned}
\label{eq.reparam}
& t_x^l=({g}_x - {p}_x^{l-1})/{p}_w^{l-1}, \\
& t_y^l=({g}_y - {p}_y^{l-1})/{p}_h^{l-1}, \\
&t_w^l=\operatorname{log}({g}_w/{p}_w^{l-1}), \\
& t_h^l=\operatorname{log}({g}_h/{p}_h^{l-1})
\end{aligned}
\end{equation}
where ${p}_x^{l-1}$/${p}_y^{l-1}$/${p}_w^{l-1}$/${p}_h^{l-1}$ are the predicted unnormalized box positions and sizes of $(l{-}1)$-th decoder layer. 

Table~\ref{tab:approach_ablation} shows that this modification can enhance the overall detection performance 
by +2.2 AP. Especially, it achieves a larger +2.9 AP improvements on small objects.

\begin{table}[t]
\renewcommand{\arraystretch}{1.5}
\centering
{\begin{center}
\tablestyle{4pt}{1.1}
\setlength{\tabcolsep}{3.75pt}
\begin{tabular}{c|c|c|cccccc}
    BoxRPB & MIM & reparam. & AP & AP$_{50}$ & AP$_{75}$ & AP$_{S}$ & AP$_{M}$ & AP$_{L}$ \\ 
    \shline
   \xmark & \xmark & \xmark & $37.2$ & $63.7$ & $37.6$ & $17.8$ & $40.5$ & $55.6$ \\
   \cmark &\xmark & \xmark & $46.1$ & $67.6$ & $49.1$ & $27.2$ & $50.5$ & $64.9$ \\
 \xmark &\cmark & \xmark & $44.6$ & $67.0$ 
 & $48.3$ & $26.9$ & $49.1$ & $59.1$ \\
 \xmark &\cmark & \cmark & $46.3$ & $68.2$ & $51.1$ & $30.7$ & $51.0$ & $58.4$ \\
\cmark &\cmark & \xmark & $48.7$ & $67.7$ & $53.0$ & $31.3$ & $53.1$ & $63.0$ \\
\cmark &\cmark & \cmark & $\bf{50.9}$ & $\bf{69.3}$ & $\bf{55.5}$ & $\bf{34.2}$ & $\bf{55.1}$ & $\bf{65.5}$ \\
\end{tabular}
\end{center}}
\caption{{\small\textbf{Core ablation results of the proposed components.} Equipped with these components, a plain DETR could achieve competitive performance. }}
\label{tab:approach_ablation}
\end{table}

\begin{table*}[t]
\centering
\subfloat[
axial decomposition.
\label{tab:box_rpb_ablation:decomp}
]{
\centering
\begin{minipage}{0.22\linewidth}{\begin{center}
\tablestyle{1pt}{1.2}
\begin{tabular}{c|c|c|ccc}
   decomp. & mem. & GFLOPs & AP & AP$_{50}$ & AP$_{75}$\\
    \shline
    \xmark & $26.8$G & $265.4$ & $50.8$ & $\bf{69.3}$ & $55.4$ \\ 
    \cmark & $9.5$G & $5.8$ & $\bf{50.9}$ & $\bf{69.3}$ & $\bf{55.5}$ \\
\end{tabular}
\end{center}}\end{minipage}
}
\hfill
\subfloat[
box points.
\label{tab:box_rpb_ablation:point}
]{
\centering
\begin{minipage}{0.12\linewidth}{\begin{center}
\tablestyle{1pt}{1.2}
\begin{tabular}{l|ccc}
   box points & AP & AP$_{50}$ & AP$_{75}$ \\
    \shline
    center & $48.0$ & $69.0$ & $53.3$\\ 
    $2\times$corners & $\bf{50.9}$ & $\bf{69.3}$ & $\bf{55.5}$  \\ 
\end{tabular}
\end{center}}\end{minipage}
}
\hfill
\subfloat[
hidden dim.
\label{tab:box_rpb_ablation:hidden_dim}
]{
\centering
\begin{minipage}{0.13\linewidth}{\begin{center}
\tablestyle{1pt}{1.2}
\begin{tabular}{l|ccc}
   hidden dim. & AP & AP$_{50}$ & AP$_{75}$ \\
    \shline
    $128$ & $50.4$ & $69.1$ & $55.1$ \\ 
    $256$ & $\bf{50.9}$ & $\bf{69.4}$ & $55.4$ \\ 
    $512$ & $\bf{50.9}$ & $69.3$ & $\bf{55.5}$ \\
\end{tabular}
\end{center}}\end{minipage}
}
\hfill
\subfloat[
cross-attention modulation.
\label{tab:box_rpb_ablation:cross_attn_modulation}
]{
\centering
\begin{minipage}{0.25\linewidth}{\begin{center}
\tablestyle{1pt}{1}
\begin{tabular}{l|ccc}
   method & AP & AP$_{50}$ & AP$_{75}$\\
    \shline
    standard cross attn. & $46.3$ & $68.2$ & $51.1$ \\ 
    conditional cross attn. & $48.3$ & $68.8$ & $52.9$ \\ 
    DAB cross attn. & $48.4$ & $68.9$ & $53.4$ \\ 
    SMCA cross attn. & $48.7$ & $69.2$ & $53.6$ \\ 
    ours  & $\bf{50.9}$ & $\bf{69.3}$ & $\bf{55.5}$ \\ 
\end{tabular}
\end{center}}\end{minipage}
}
\caption{{\small\textbf{Ablation of box relative position bias scheme.} (a) Axial decomposition can significantly decrease the computation overhead and GPU memory footprint. (b) The corner points perform better than the center point. (c) The higher the hidden dimension, the better performance. (d) Our approach performs much better than other related methods designed to modulate the cross-attention maps}.}
\label{tab:box_rpb_ablation}
\vspace{-3mm}
\end{table*}

\begin{table}[t]
\renewcommand{\arraystretch}{1.5}
\centering
{\begin{center}
\tablestyle{4pt}{1.25}
\begin{tabular}{l|cccccc}
    method & AP & AP$_{50}$ & AP$_{75}$ & AP$_{S}$ & AP$_{M}$ & AP$_{L}$ \\
    \shline
    deformable cross attn. & $50.2$ & $68.5$ & $54.8$ & $34.1$ & $54.4$ & $63.3$ \\ 
    RoIAlign & $49.6$ & $68.3$ & $54.1$ & $31.9$ & $54.2$ & $63.5$  \\
    RoI Sampling & $49.3$ &  $68.2$ & $53.8$ &$33.1$ & $53.2$ & $63.0$ \\
    Box Mask & $48.6$ &  $68.7$ & $52.9$ &$31.8$ & $52.7$ & $63.0$ \\
    Ours & $\bf{50.9}$ & $\bf{69.3}$ & $\bf{55.5}$ & $\bf{34.2}$ & $\bf{55.1}$ & $\bf{65.5}$ \\
\end{tabular}
\end{center}}
\caption{{\small\textbf{Comparison with local cross-attention scheme.} Global cross-attention with BoxRPB outperforms all the local cross-attention counterparts and have a significant gain on large objects.
}}
\label{tab:comp_to_local}
\end{table}

\vspace{1mm}
\section{Ablation Study and Analysis}\label{sec:empirical_study}

\vspace{1mm}
\subsection{The importance of box relative position bias}\label{sec:exp_box_rpb}
In Table~\ref{tab:box_rpb_ablation}, we study the effect of each factor within our BoxRPB scheme and report the detailed comparison results in the following discussion.

\vspace{1mm}
\noindent\textbf{Effect of axial decomposition.}
Modeling the 2D relative position without any decomposition is a naive baseline compared with our axial decomposition schema, and it can be parameterized as $(\Delta\mathbf{x}_1, \Delta\mathbf{y}_1, \Delta\mathbf{x}_2, \Delta\mathbf{y}_2)\in \mathbb{R}^{\mathsf{K}\times\mathsf{H}\times\mathsf{W}\times4} $. This baseline requires a quadratic computation overhead and memory consumption while the decomposed one decreases the cost to linear complexity. In Table~\ref{tab:box_rpb_ablation:decomp}, we compared the two approaches and find that the axial decomposition scheme achieves comparable performance ($50.9$ vs. $50.8$) while it requires a much lower memory footprint ($9.5$G vs. $26.8$G) and smaller computation overhead ($5.8$G FLOPs vs. $265.4$G FLOPs).

\vspace{1mm}
\noindent\textbf{Effect of box points.}
Table~\ref{tab:box_rpb_ablation:point} shows the comparison of using only the center points or the two corner points.
We find that applying only the center points improves the baseline (fourth row of Table~\ref{tab:approach_ablation}) by +1.7 AP. However, its performance is worse than that of using two corner points. In particular, while the two methods achieve comparable AP$_{50}$ results, utilizing corner points could boost AP$_{75}$ by +2.2.
This shows that not only the position (center) but also the scale (height and width) of the query box are important to precisely model relative position bias.

\vspace{1mm}
\noindent\textbf{Effect of hidden dimension.}
We study the effect of the hidden dimension in Equation~\ref{eq.rpe_mlp}. As shown in Table~\ref{tab:box_rpb_ablation:hidden_dim}, a smaller hidden dimension of 128 would lead to a performance drop of 0.5, indicating that the position relation is non-trivial and requires a higher dimension space to model.

\vspace{1mm}
\noindent\textbf{Comparison with other methods.}
We study the effect of choosing other schemes to compute the modulation term $\mathbf{B}$ in Equation~\ref{eq.general_cross_attn}. We compared to several representative methods as follows:
(i) Conditional cross-attention scheme~\cite{meng2021CondDETR}, which computes the modulation term based on the inner product between the conditional spatial (position) query embedding and the spatial key embedding.
(ii) DAB cross-attention scheme~\cite{liu2022dab}, which builds on conditional cross-attention and further modulates the positional attention map using the box width
and height information.
(iii) Spatially modulated cross-attention scheme (SMCA)~\cite{gao2021fast}, which designs handcrafted query spatial priors, implemented with a 2D Gaussian-like weight map, to constrain the attended features to be around the
object queries’ initial estimations.

Table~\ref{tab:box_rpb_ablation:cross_attn_modulation} reports the detailed comparison results. Our approach achieves the best performance among all the methods.
Specifically, the conditional cross-attention module achieves similar performance with our center-only setting (first row of Table~\ref{tab:box_rpb_ablation:point}).
DAB cross-attention and SMCA are slightly better than the conditional cross-attention module, but they still lag behind the BoxRPB by a gap of 2.5 AP and 2.2 AP, respectively.

We also compare BoxRPB with DAB cross-attention based on its official open-source code. Replacing DAB positional module with BoxRPB achieves a +1.8 mAP performance gain.

\begin{figure*}[t]
\centering
\begin{subfigure}[b]{0.3\linewidth}
\includegraphics[width=0.99\textwidth]{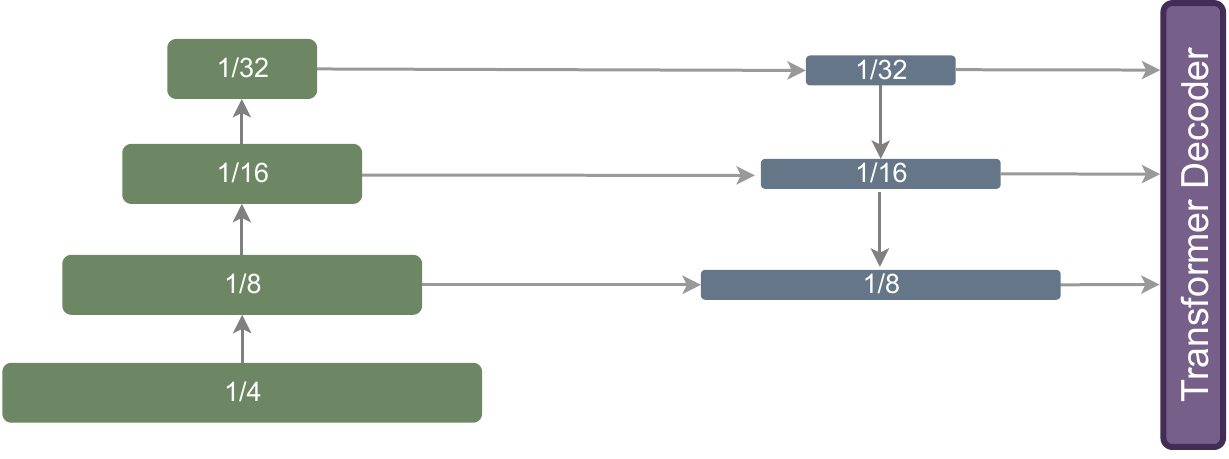}
\caption{\small{$(\mathbf{C}^{3}$,$\mathbf{C}^{4}$,$\mathbf{C}^{{5}}$) $\to$  $(\mathbf{P}^{{3}}$, $\mathbf{P}^{{4}}$, $\mathbf{P}^{{5}})$}}
\end{subfigure}
\hfill
\begin{subfigure}[b]{0.3\linewidth}
\includegraphics[width=0.99\textwidth]{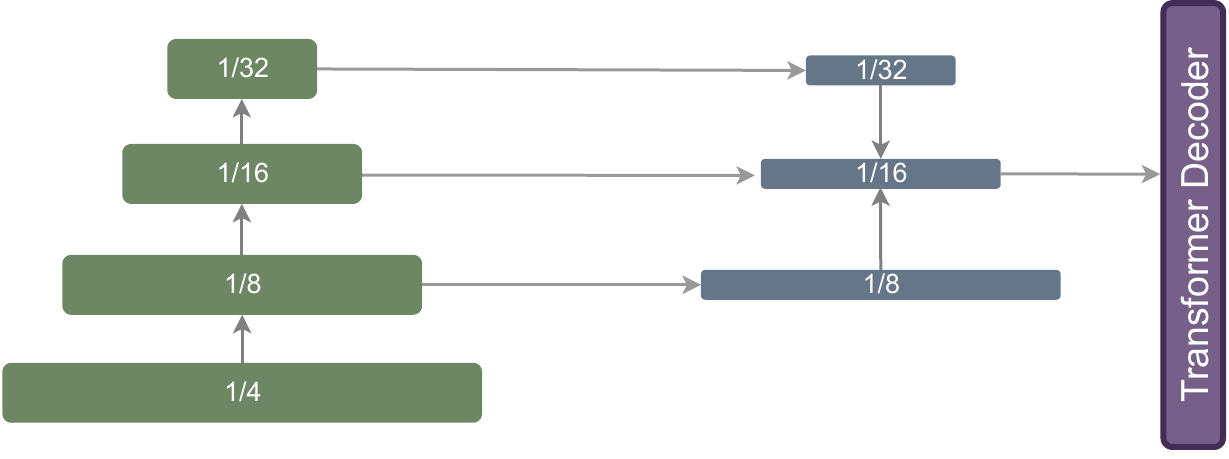}
\caption{\small{$(\mathbf{C}^{3}$,$\mathbf{C}^{4}$,$\mathbf{C}^{{5}}$) $\to$ $\mathbf{P}^{4}$}}
\end{subfigure}%
\hfill
\begin{subfigure}[b]{0.3\linewidth}
\includegraphics[width=0.99\textwidth]{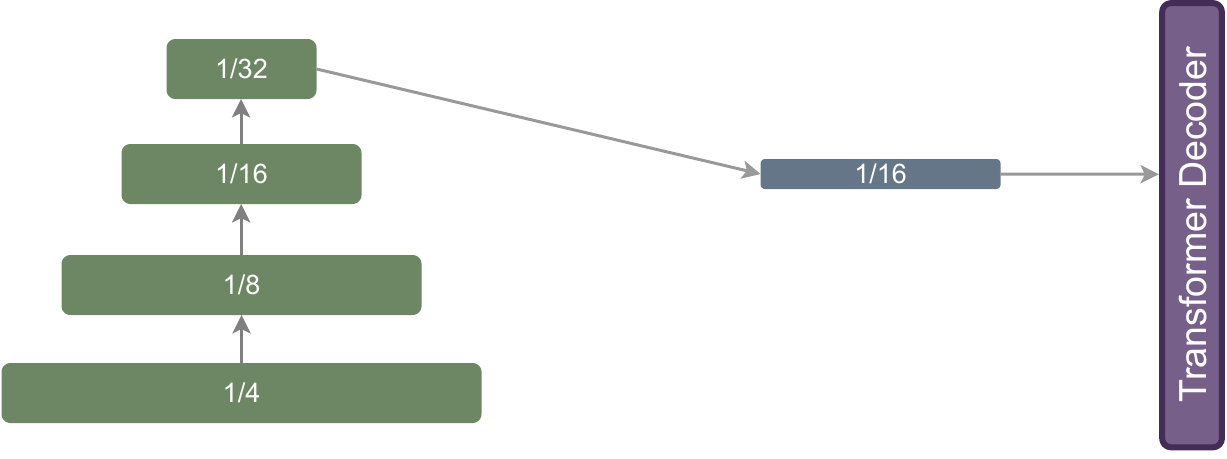}
\caption{\small{$\mathbf{C}^{{5}}$ $\to$ $\mathbf{P}^{4}$}}
\end{subfigure}%
\caption{\small{
We compare the architecture designs when using different feature maps output by the backbone and sent to the Transformer decoder.
From (a) to (b), we simplify the dependency on sending multi-scale feature maps to the Transformer decoder.
From (b) to (c), we remove the dependency on fusing multi-scale feature output by the backbone.
We adopt (c) as our default architecture setting.
}}
\label{fig:ms_feat}
\end{figure*}

\subsection{Comparison with local attention scheme}\label{sec:exp_local_attn}
In this section, we compared our global attention schema with other representative local cross-attention mechanisms, including 
deformable cross-attention ~\cite{zhu2020deformable}, RoIAlign~\cite{he2017mask}, RoI Sampling (sampling fixed points inside the Region of Interest), and box mask inspired by~\cite{cheng2021masked}. We illustrate the key differences between those methods in 
the supplementary material.

As shown in Table~\ref{tab:comp_to_local}, our method surpasses all the local cross-attention variants. In addition, we observed that large objects have larger improvements for our method. A similar observation is also reported in DETR~\cite{carion2020end}, it may be due to more effective long-range context modeling based on the global attention scheme.

\begin{table}[t]
\centering
\subfloat[
\label{tab:mim_ablation:1}
]{
\centering
\begin{minipage}{\linewidth}{\begin{center}
\tablestyle{1pt}{1.2}
\begin{tabular}{l|c|cccccc}
   backbone $\to$ decoder & 
   MIM & AP & AP$_{50}$ & AP$_{75}$ & AP$_{S}$ & AP$_{M}$ & AP$_{L}$ \\
    \shline
   $(\mathbf{C}^{3}$,$\mathbf{C}^{4}$,$\mathbf{C}^{{5}}$) $\to$  $(\mathbf{P}^{{3}}$, $\mathbf{P}^{{4}}$, $\mathbf{P}^{{5}})$ 
     & \xmark & $49.6$ & $69.2$ & $53.8$ & $31.5$ & $53.4$ & $\bf{65.2}$ \\ 
   $(\mathbf{C}^{3}$,$\mathbf{C}^{4}$,$\mathbf{C}^{{5}}$) $\to$  $(\mathbf{P}^{{3}}$, $\mathbf{P}^{{4}}$, $\mathbf{P}^{{5}})$ 
     & \cmark & $\bf{51.1}$ & $\bf{69.3}$ & $\bf{56.0}$ & $\bf{34.8}$ & $\bf{55.4}$ & $\bf{65.2}$ \\ 
\end{tabular}
\end{center}}
\end{minipage}
}
\hfill
\subfloat[
\label{tab:mim_ablation:2}
]{
\centering
\begin{minipage}{\linewidth}{\begin{center}
\tablestyle{3pt}{1.2}
\resizebox{1\linewidth}{!}
{
\begin{tabular}{l|c|cccccc}
   backbone $\to$ decoder &
   MIM & AP & AP$_{50}$ & AP$_{75}$ & AP$_{S}$ & AP$_{M}$ & AP$_{L}$ \\
    \shline
    $(\mathbf{C}^{3}$,$\mathbf{C}^{4}$,$\mathbf{C}^{{5}}$) $\to$ $\mathbf{P}^{5}$ & \xmark &  $47.0$ & $68.2$ & $50.4$ & $28.0$ & $51.5$ & $64.2$ \\ 
    $(\mathbf{C}^{3}$,$\mathbf{C}^{4}$,$\mathbf{C}^{{5}}$) $\to$ $\mathbf{P}^{4}$ & \xmark &  $49.6$ & $\bf{69.8}$ & $53.4$ & $31.4$ & $53.7$  & $\bf{65.5}$ \\ 
    $(\mathbf{C}^{3}$,$\mathbf{C}^{4}$,$\mathbf{C}^{{5}}$) $\to$ $\mathbf{P}^{3}$ & \xmark &  $49.7$ & $\bf{69.8}$ & $53.9$ & $32.7$ & $53.5$ & $65.2$ \\ 
    $(\mathbf{C}^{3}$,$\mathbf{C}^{4}$,$\mathbf{C}^{{5}}$) $\to$ $\mathbf{P}^{5}$ & \cmark &  $50.3$ & $69.3$ & $54.9$ & $33.4$ & $54.7$ & $64.9$ \\ 
    $(\mathbf{C}^{3}$,$\mathbf{C}^{4}$,$\mathbf{C}^{{5}}$) $\to$ $\mathbf{P}^{4}$ & \cmark &  $\bf{51.0}$ & $69.4$ & $\bf{55.7}$ & $\bf{34.5}$ & $\bf{55.1}$ &  $65.2$\\
    $(\mathbf{C}^{3}$,$\mathbf{C}^{4}$,$\mathbf{C}^{{5}}$) $\to$ $\mathbf{P}^{3}$ & \cmark &  $50.9$ & $69.2$ & $55.4$ & $34.4$ & $55.0$ & $64.5$ \\
\end{tabular}
}
\end{center}}
\end{minipage}
}
\hfill
\subfloat[
\label{tab:mim_ablation:3}
]{
\centering
\begin{minipage}{\linewidth}{\begin{center}
\tablestyle{3pt}{1.2}
\resizebox{1\linewidth}{!}
{
\begin{tabular}{l|c|cccccc}
   backbone $\to$ decoder &
   MIM & AP & AP$_{50}$ & AP$_{75}$ & AP$_{S}$ & AP$_{M}$ & AP$_{L}$ \\
    \shline
    $\mathbf{C}^{{5}}$ $\to$ $\mathbf{P}^{5}$ & \xmark & $46.4$ & $67.7$ & $49.7$ & $26.9$ & $50.5$ & $64.4$ \\ 
    $\mathbf{C}^{{5}}$ $\to$ $\mathbf{P}^{4}$ & \xmark & $48.0$ & $68.7$ & $51.8$ & $30.4$ & $52.2$ & $64.4$ \\
    $\mathbf{C}^{{5}}$ $\to$ $\mathbf{P}^{3}$ & \xmark & $48.7$ & $69.1$ & $52.6$ & $30.7$ & $52.9$ & $64.9$ \\
    $\mathbf{C}^{{5}}$ $\to$ $\mathbf{P}^{5}$ & \cmark & $50.2$ & $69.1$ & $55.0$ & $33.5$ & $54.5$ & $64.6$ \\ 
    $\mathbf{C}^{{5}}$ $\to$ $\mathbf{P}^{4}$ & \cmark & $\bf{50.9}$ & $\bf{69.3}$ & $55.5$ & $34.2$ & $\bf{55.1}$ & $\bf{65.5}$ \\
    $\mathbf{C}^{{5}}$ $\to$ $\mathbf{P}^{3}$ & \cmark & $\bf{50.9}$ & $69.2$ & $\bf{55.7}$ & $\bf{34.6}$ & $54.9$ & $65.0$ \\
\end{tabular}
}
\end{center}}
\end{minipage}
}
\caption{{\small\textbf{Ablation of MIM pre-training.} (a) multi-scale feature maps output by the backbone + multi-scale feature maps for the Transformer decoder. (b) multi-scale feature maps output by the backbone + single-scale feature map for the Transformer decoder. (c) single-scale feature map output by the backbone + single-scale feature map for the Transformer decoder.}}
\label{tab:mim_ablation}
\end{table}

\vspace{1mm}
\subsection{On MIM pre-training}\label{sec:exp_mim_pretrain}
We explore different ways of using the backbone and decoder feature maps with or without MIM pre-training. We evaluate the performance of three different architecture configurations, which are illustrated in Figure~\ref{fig:ms_feat}. We discuss and analyze the results as follows.

\vspace{1mm}
\noindent\textbf{MIM pre-training brings consistent gains.}
By comparing the experimental results under the same architecture configuration, we found that using MIM pre-training consistently achieves better performance. 
For example, as shown in Table~\ref{tab:mim_ablation}, using MIM pre-training outperforms using supervised pre-training by 1.5 AP in the$(\mathbf{C}^{3}$,$\mathbf{C}^{4}$,$\mathbf{C}^{{5}}$) $\to$  $(\mathbf{P}^{{3}}$, $\mathbf{P}^{{4}}$, $\mathbf{P}^{{5}})$ configuration and 2.9 AP in the $\mathbf{C}^{{5}}$ $\to$  $\mathbf{P}^{{4}}$ configuration.

\vspace{1mm}
\noindent\textbf{Multi-scale feature maps for the decoder can be removed.} By comparing the results between Table~\ref{tab:mim_ablation:1} and Table~\ref{tab:mim_ablation:2}, we found that using high-resolution feature maps can match or even surpass the performance of using multi-scale feature maps. For example, 
($\mathbf{C}^{3}$,$\mathbf{C}^{4}$,$\mathbf{C}^{{5}}$) $\to$  $\mathbf{P}^{{3}}$ achieves comparable performance with ($\mathbf{C}^{3}$,$\mathbf{C}^{4}$,$\mathbf{C}^{{5}}$) $\to$  $(\mathbf{P}^{{3}}$, $\mathbf{P}^{{4}}$, $\mathbf{P}^{{5}})$ with or without using MIM pre-training. This observation is not trivial as most existing detection heads still require multi-scale features as input, and it makes building a competitive single-scale plain DETR possible. We hope this finding could ease the design of future detection frameworks.

\noindent\textbf{Multi-scale feature maps from the backbone are non-necessary.} 
We analyze the effect of removing the multi-scale feature maps from the backbone by comparing the results of Table~\ref{tab:mim_ablation:2} and Table~\ref{tab:mim_ablation:3}. When using a supervised pre-trained backbone, adopting only the last feature map $\mathbf{C}^{{5}}$ from the backbone would hurt the performance. For example, when using the supervised pre-trained backbone, the
$\mathbf{C}^{5}$ $\to$ $\mathbf{P}^{5}$ reaches 46.4 AP, which is worse than ($\mathbf{C}^{3}$,$\mathbf{C}^{4}$,$\mathbf{C}^{{5}}$) $\to$ $\mathbf{P}^{5}$ (47.0 AP) by 0.6 AP. However, when using the MIM pre-trained backbone, $\mathbf{C}^{5}$ $\to$ $\mathbf{P}^{5}$ reaches 50.2 mAP, which is comparable with the performance of ($\mathbf{C}^{3}$,$\mathbf{C}^{4}$,$\mathbf{C}^{{5}}$) $\to$ $\mathbf{P}^{5}$ (50.3 AP). 
These results show that MIM pre-training can reduce the reliance on multi-scale feature maps.

\vspace{1mm}
\noindent\textbf{Single-scale feature map from the backbone and single-scale feature map for the decoder is enough.}
Based on the above observations, we can reach a surprisingly simple but important conclusion that we can completely eliminate the need for multi-scale feature maps in both the backbone and Transformer decoder by using our proposed BoxRPB scheme and MIM pre-training.

\vspace{1mm}
\begin{table}[t]
\renewcommand{\arraystretch}{1.5}
\centering
{\begin{center}
\tablestyle{4pt}{1.25}
\begin{tabular}{l|cccccc}
    method & AP & AP$_{50}$ & AP$_{75}$ & AP$_{S}$ & AP$_{M}$ & AP$_{L}$ \\
    \shline
    Cascade Mask R-CNN\cite{cai2018cascade} & $53.7$ & $71.9$ & $58.7$ & $\bf{36.9}$ & $\bf{57.4}$ & $\bf{69.1}$ \\
    Ours & $\bf{53.8}$ & $\bf{73.4}$ & $\bf{58.9}$ & $35.9$ & $57.0$ & $68.9$ \\
\end{tabular}
\end{center}}
\caption{{\small\textbf{Comparison of the improved plain DETR and Cascade Mask R-CNN with a MIM pre-trained ViT-Base backbone.} Our plain DETR with global cross-attention is slightly better than the region-based, multi-scaled Cascade Mask R-CNN.
}}
\label{tab:comp_with_vit}
\end{table}

\vspace{1mm}
\subsection{Application to a plain ViT}\label{sec:vit_exps}
In this section, we build a simple and effective fully plain object detection system by applying our approach to the plain ViT~\cite{vit20}. Our system only uses a single-resolution feature map throughout a plain Transformer encoder-decoder architecture, without any multi-scale design or processing. We compare our approach with the state-of-the-art Cascade Mask R-CNN~\cite{cai2018cascade, li2022exploring} on the COCO dataset. For the fair comparison, We use a MAE~\cite{he2022masked} pre-trained ViT-Base as the backbone and train the object detector for {$\sim$}$50$ epochs. As shown in Table~\ref{tab:comp_with_vit}, our method achieves comparable results with Cascade Mask R-CNN which relies on using multi-scale feature maps for better localization across different object scales. Remarkably, our method does not train with instance mask annotations that are usually considered to be beneficial for object detection.

\subsection{Visualization of cross-attention maps}\label{sec:visualization}

\begin{figure}[]
    \centering
    \begin{minipage}[t]{0.475\textwidth}
    \begin{center}
    \includegraphics[width=0.99\textwidth]{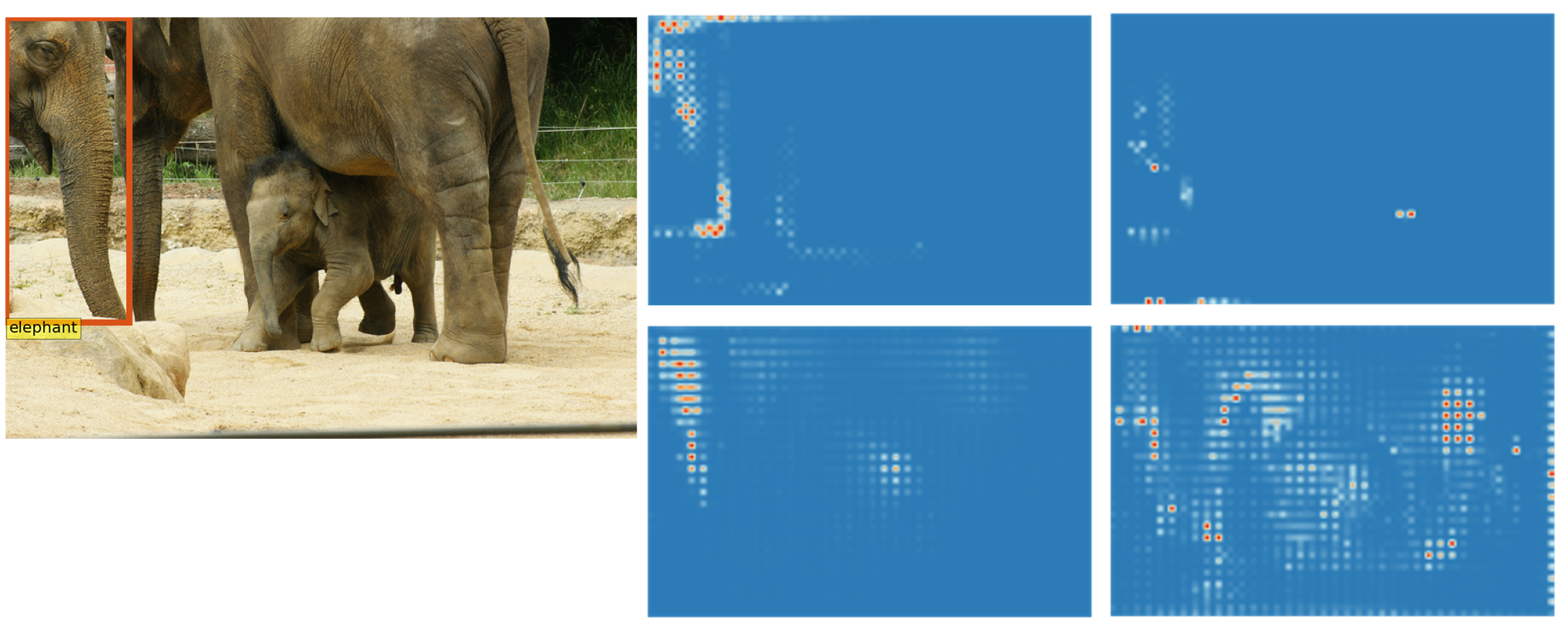}
    \end{center}
    \end{minipage}
    \\
    \begin{minipage}[t]{0.475\textwidth}
    \begin{center}
    \includegraphics[width=0.99\textwidth]{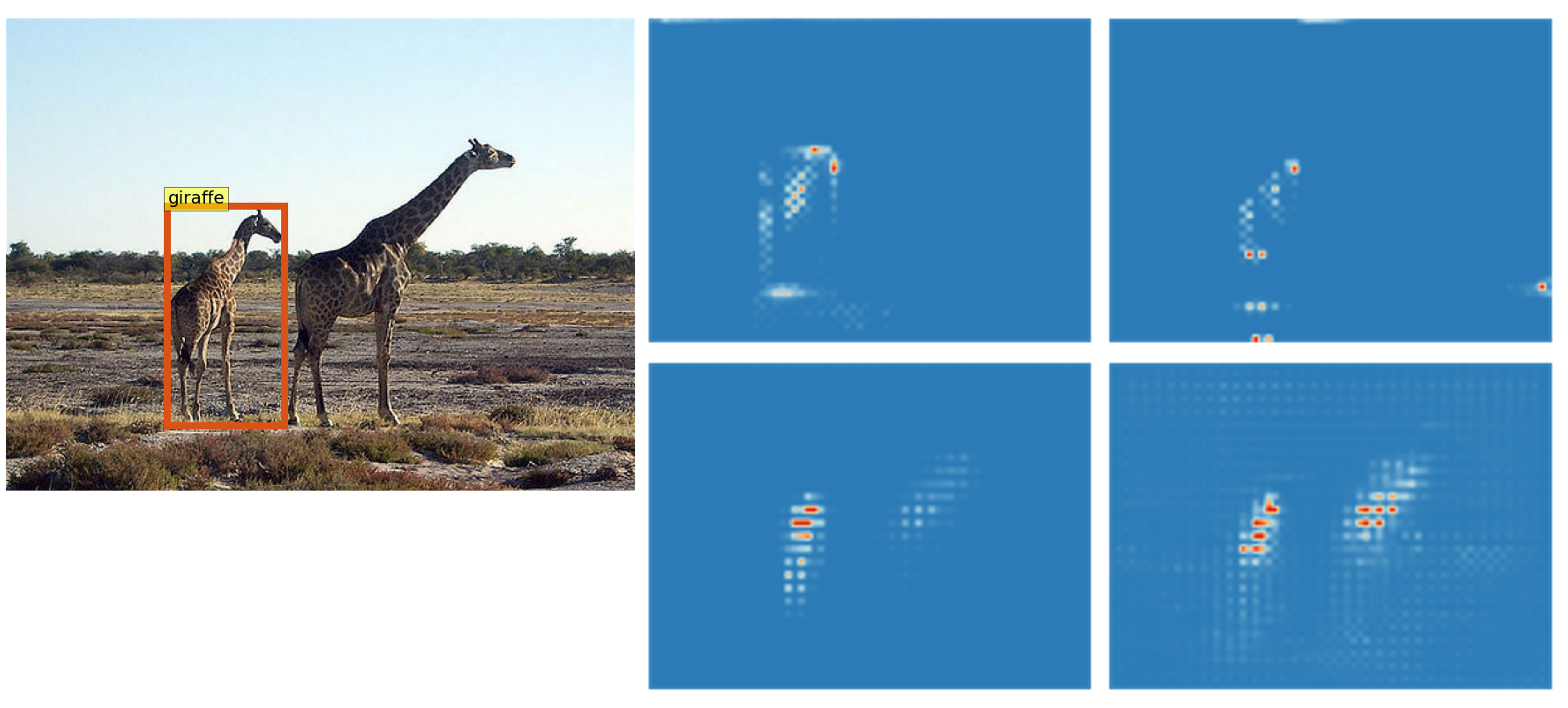}
    \end{center}
    \end{minipage}
    \\
    \caption{\small{Visualizations of the cross-attention maps of models w. or w/o. BoxRPB.} For each group, the first column shows the input image and the object query. The first row presents the attention maps of the model w. BoxRPB, while the second row displays attention maps of the model w/o. BoxRPB. BoxRPB helps to guide the cross-attention to focus on the individual objects.}
    \label{fig:cross_attn}
\end{figure}

Figure~\ref{fig:cross_attn} shows the cross-attention maps of models with or without BoxRPB.
For the model with BoxRPB, the cross-attention concentrate on the individual object. In the contrary, the cross-attention of model without BoxRPB attend to multiple objects that have similar appearance.

\begin{table*}[t]
\begin{minipage}[t]{1\linewidth}
\newcommand{\blue}[1]{\textcolor{blue}{#1}}
    \definecolor{deepgreen}{rgb}{0.07, 0.53, 0.03}
    \centering
    \setlength{\tabcolsep}{8pt}
    \renewcommand{\arraystretch}{1.5}
    \footnotesize
    \resizebox{0.99\linewidth}{!}
    {
        \begin{tabular}{l|c|c|c|c|cccccc}
            method     & framework & extra data & \#params & \#epoch & AP & AP$_{50}$ & AP$_{75}$ & AP$_{S}$ & AP$_{M}$ & AP$_{L}$ \\
            \shline
            Swin ~\cite{liu2021swin} & HTC & & 284M& $72$  &  $57.7$ & $76.2$ & $63.1$ & $33.4$ & $52.9$ & $64.0$  \\
            DETA~\cite{ouyangzhang2022nms} & DETR & & 218M & $24$ & $58.5$ & $76.5$ & $64.4$ & $38.5$ & $62.6$ & $73.8$ \\
            DINO-DETR~\cite{zhang2022dino} & DETR & & 218M  & $36$ & $58.6$ & $76.9$ & $64.1$ & $39.4$ & $61.6$ & $73.2$ \\
            Ours$^*$ & DETR &  & 228M & $36$ & $60.0$ & $78.9$ & $66.4$ &  $42.8$ & $62.7$ & $73.7$ \\
            \hline
            DETA~\cite{ouyangzhang2022nms} & DETR & O365 & 218M & $24+24$ & $63.5$ & $80.4$ & $70.2$ & $46.1$ & $\bf{66.9}$ & $\bf{76.9}$  \\
            DINO-DETR~\cite{zhang2022dino}$^*$  & DETR & O365 & 218M  & $26+18$ & {$63.3$} & {$-$} & {$-$} & {$-$} & {$-$} & {$-$} \\
            Ours$^*$ & DETR & O365 & 228M & $24+24$ & $\bf{63.9}$ & $\bf{82.1}$ & $\bf{70.7}$ & $\bf{48.2}$ & $66.8$ & $76.7$ \\
        \end{tabular}
    }
    \caption{\small{System-level comparisons with the state-of-the-art results on COCO \texttt{test-dev}. All methods adopt the Swin-Large backbone. The $^*$ marks the results with test time augmentation. }
    \label{tab:compare_to_sota_coco}
    }
\end{minipage}
\end{table*}

\section{System-level Results}

We compare our method with other state-of-the-art methods in this section. Table~\ref{tab:compare_to_sota_coco} shows results, where all experiments reported in this table utilize a Swin-Large as the backbone. As other works usually apply an encoder to enhance the backbone features, we also stack 12 window-based single-scale transformer layers (with a feature dimension of 256) on top of the backbone for a fair comparison.

With the 36 training epochs, our model achieves 
$60.0$ AP on the COCO test-dev set, which outperforms DINO-DETR by 1.4 AP. Further introducing the Objects365~\cite{shao2019objects365} as the pre-training dataset, our method reaches 
$63.9$ AP on the test-dev set, which is better than DINO-DETR and DETA by a notable margin. These strong results verify that the plain DETR architecture does not have intrinsic drawbacks to prevent it from achieving high performance.

\section{Related work}

\vspace{1mm}
\noindent\textbf{DETR-based object detection.}
DETR~\cite{carion2020end} has impressed the field for its several merits, including the conceptually straightforward and generic in applicability, requiring minimal domain knowledge that avoids customized label assignments and non-maximum suppression, and being plain. While the original DETR maintains a plain design, it also suffers from slow convergence rate and lower detection accuracy. There have been many follow-up works including~\cite{meng2021CondDETR,gao2021fast,dai2021dynamic,wang2021anchor,zhu2020deformable,zhang2022towards,zhang2022semantic,gao2022adamixer,zhang2022dino}, and now many top object detectors have been built upon this line of works, thanks to the reintroduction of multi-scale and locality designs~\cite{zhang2022dino,fang2022eva,wang2022internimage}. Unlike these leading works, we aim for an improved DETR framework that maintains a ``plain'' nature without multi-scale features and local cross-attention computation.

\vspace{3mm}
\noindent\textbf{Region-based object detection.}
Prior to the DETR framework, the object detectors were usually built in a region-based fashion: the algorithms analyze every region of the entire image locally, and the object detections are obtained by ranking and filtering the results of each region. Due to the locality nature, it's hard for them to flexibly leverage global information for object detection. Moreover, while some early attempts use single scale feature map on the head~\cite{girshick2014rich,Redmon_2016_CVPR,girshick2015fast,ren2015faster,liu2016ssd}, later, the leading methods are almost all built by multi-scale features such as FPN~\cite{lin2017feature}, BiFPN~\cite{tan2020efficientdet}, Cascade R-CNN~\cite{cai2018cascade}, and HTC~\cite{chen2019hybrid}, etc. We expect our strong plain DETR detector may also inspire research in exploring single-scale feature map for region-based detection.

\vspace{3mm}
\noindent\textbf{Position encoding.} This paper is also related to position encoding techniques. The original Transformer~\cite{vaswani2017attention} uses absolute position encoding. Early vision Transformers~\cite{carion2020end,dosovitskiy2020image,touvron2021training} inherit this absolute position encoding setting. Swin Transformers~\cite{liu2021swin,liu2022swin} highlight the importance of relative position bias for Transformer-based visual recognition, where some early variants can be found in both language and vision domains~\cite{hu2018relation,shaw2018self,hu2019local,dai2019transformer,huang2020improve,chu2021we,wu2021rethinking}. This paper extends the relative position bias for box-to-pixel pairs, instead of previous pixel-to-pixel pairs. It also reveals that the RPB can effect even more critical in the context of plain DETR detectors.

\vspace{3mm}
\noindent\textbf{Pre-training.}
The pre-training methods~\cite{he2022masked,xie2022simmim,bao2021beit} that follow the path of masked image modeling have drawn increasing attention due to their strong performance on various core vision tasks such as object detection and semantic segmentation. Although some recent works~\cite{li2022exploring,xie2022revealing} have revealed some possible reasons why MIM outperforms the conventional supervised pre-training and confirmed that FPN can be simplified, few works attempt to build a fully plain object detection head based on MIM pre-trained backbones.
Our experiment results show that MIM pre-training is a key factor in fully plain object detection architecture design.

\vspace{2mm}
\section{Conclusion}
This paper has present an improved plain DETR detector which achieves exceptional improvements over the original plain model, and achieves a 63.9 mAP accuracy using a Swin-L backbone, which is highly competitive with state-of-the-art detectors that have been heavily tuned using multi-scale feature maps and region-based feature extraction. We highlighted the importance of two technologies of BoxRPB and MIM-based pre-training for this improved plain DETR framework. We hope the effective detector empowered by minimal architectural ``inductive bias'' can encourage future research to explore generic plain decoders in other vision problems.

{\small
\bibliographystyle{ieee}
\bibliography{egbib}

\begin{thebibliography}{10}\itemsep=-1pt

\bibitem{bao2021beit}
H.~Bao, L.~Dong, S.~Piao, and F.~Wei.
\newblock Beit: Bert pre-training of image transformers.
\newblock {\em arXiv preprint arXiv:2106.08254}, 2021.

\bibitem{brown2020language}
T.~Brown, B.~Mann, N.~Ryder, M.~Subbiah, J.~D. Kaplan, P.~Dhariwal,
  A.~Neelakantan, P.~Shyam, G.~Sastry, A.~Askell, et~al.
\newblock Language models are few-shot learners.
\newblock {\em Advances in neural information processing systems},
  33:1877--1901, 2020.

\bibitem{cai2018cascade}
Z.~Cai and N.~Vasconcelos.
\newblock Cascade r-cnn: Delving into high quality object detection.
\newblock In {\em CVPR}, pages 6154--6162, 2018.

\bibitem{carion2020end}
N.~Carion, F.~Massa, G.~Synnaeve, N.~Usunier, A.~Kirillov, and S.~Zagoruyko.
\newblock End-to-end object detection with transformers.
\newblock In {\em European conference on computer vision}, pages 213--229.
  Springer, 2020.

\bibitem{chen2019hybrid}
K.~Chen, J.~Pang, J.~Wang, Y.~Xiong, X.~Li, S.~Sun, W.~Feng, Z.~Liu, J.~Shi,
  W.~Ouyang, et~al.
\newblock Hybrid task cascade for instance segmentation.
\newblock In {\em CVPR}, pages 4974--4983, 2019.

\bibitem{chen2022group}
Q.~Chen, J.~Wang, C.~Han, S.~Zhang, Z.~Li, X.~Chen, J.~Chen, X.~Wang, S.~Han,
  G.~Zhang, et~al.
\newblock Group detr v2: Strong object detector with encoder-decoder
  pretraining.
\newblock {\em arXiv preprint arXiv:2211.03594}, 2022.

\bibitem{cheng2021masked}
B.~Cheng, I.~Misra, A.~G. Schwing, A.~Kirillov, and R.~Girdhar.
\newblock Masked-attention mask transformer for universal image segmentation.
\newblock {\em arXiv preprint arXiv:2112.01527}, 2021.

\bibitem{chu2021we}
X.~Chu, B.~Zhang, Z.~Tian, X.~Wei, and H.~Xia.
\newblock Do we really need explicit position encodings for vision
  transformers.
\newblock {\em arXiv preprint arXiv:2102.10882}, 3(8), 2021.

\bibitem{dai2021dynamic}
X.~Dai, Y.~Chen, J.~Yang, P.~Zhang, L.~Yuan, and L.~Zhang.
\newblock Dynamic detr: End-to-end object detection with dynamic attention.
\newblock In {\em Proceedings of the IEEE/CVF International Conference on
  Computer Vision}, pages 2988--2997, 2021.

\bibitem{dai2019transformer}
Z.~Dai, Z.~Yang, Y.~Yang, J.~Carbonell, Q.~V. Le, and R.~Salakhutdinov.
\newblock Transformer-xl: Attentive language models beyond a fixed-length
  context.
\newblock {\em arXiv preprint arXiv:1901.02860}, 2019.

\bibitem{devlin2018bert}
J.~Devlin, M.-W. Chang, K.~Lee, and K.~Toutanova.
\newblock Bert: Pre-training of deep bidirectional transformers for language
  understanding.
\newblock {\em arXiv preprint arXiv:1810.04805}, 2018.

\bibitem{dosovitskiy2020image}
A.~Dosovitskiy, L.~Beyer, A.~Kolesnikov, D.~Weissenborn, X.~Zhai,
  T.~Unterthiner, M.~Dehghani, M.~Minderer, G.~Heigold, S.~Gelly, et~al.
\newblock An image is worth 16x16 words: Transformers for image recognition at
  scale.
\newblock {\em arXiv preprint arXiv:2010.11929}, 2020.

\bibitem{vit20}
A.~Dosovitskiy, L.~Beyer, A.~Kolesnikov, D.~Weissenborn, X.~Zhai,
  T.~Unterthiner, M.~Dehghani, M.~Minderer, G.~Heigold, S.~Gelly, J.~Uszkoreit,
  and N.~Houlsby.
\newblock An image is worth 16x16 words: Transformers for image recognition at
  scale.
\newblock In {\em ICLR}, 2021.

\bibitem{fang2022eva}
Y.~Fang, W.~Wang, B.~Xie, Q.~Sun, L.~Wu, X.~Wang, T.~Huang, X.~Wang, and
  Y.~Cao.
\newblock Eva: Exploring the limits of masked visual representation learning at
  scale.
\newblock {\em arXiv preprint arXiv:2211.07636}, 2022.

\bibitem{MIMDet}
Y.~Fang, S.~Yang, S.~Wang, Y.~Ge, Y.~Shan, and X.~Wang.
\newblock Unleashing vanilla vision transformer with masked image modeling for
  object detection.
\newblock {\em arXiv preprint arXiv:2204.02964}, 2022.

\bibitem{gao2021fast}
P.~Gao, M.~Zheng, X.~Wang, J.~Dai, and H.~Li.
\newblock Fast convergence of detr with spatially modulated co-attention.
\newblock In {\em ICCV}, pages 3621--3630, 2021.

\bibitem{gao2022adamixer}
Z.~Gao, L.~Wang, B.~Han, and S.~Guo.
\newblock Adamixer: A fast-converging query-based object detector.
\newblock In {\em Proceedings of the IEEE/CVF Conference on Computer Vision and
  Pattern Recognition}, pages 5364--5373, 2022.

\bibitem{girshick2015fast}
R.~Girshick.
\newblock Fast r-cnn.
\newblock In {\em ICCV}, pages 1440--1448, 2015.

\bibitem{girshick2014rich}
R.~Girshick, J.~Donahue, T.~Darrell, and J.~Malik.
\newblock Rich feature hierarchies for accurate object detection and semantic
  segmentation.
\newblock In {\em Proceedings of the IEEE conference on computer vision and
  pattern recognition}, pages 580--587, 2014.

\bibitem{he2022masked}
K.~He, X.~Chen, S.~Xie, Y.~Li, P.~Doll{\'a}r, and R.~Girshick.
\newblock Masked autoencoders are scalable vision learners.
\newblock In {\em Proceedings of the IEEE/CVF Conference on Computer Vision and
  Pattern Recognition}, pages 16000--16009, 2022.

\bibitem{he2017mask}
K.~He, G.~Gkioxari, P.~Doll{\'a}r, and R.~Girshick.
\newblock {Mask R-CNN}.
\newblock In {\em ICCV}, 2017.

\bibitem{He2016ResNet}
K.~{He}, X.~{Zhang}, S.~{Ren}, and J.~{Sun}.
\newblock {Deep Residual Learning for Image Recognition}.
\newblock In {\em CVPR}, 2016.

\bibitem{hu2018relation}
H.~Hu, J.~Gu, Z.~Zhang, J.~Dai, and Y.~Wei.
\newblock Relation networks for object detection.
\newblock In {\em Proceedings of the IEEE conference on computer vision and
  pattern recognition}, pages 3588--3597, 2018.

\bibitem{hu2019local}
H.~Hu, Z.~Zhang, Z.~Xie, and S.~Lin.
\newblock Local relation networks for image recognition.
\newblock In {\em Proceedings of the IEEE/CVF International Conference on
  Computer Vision}, pages 3464--3473, 2019.

\bibitem{huang2020improve}
Z.~Huang, D.~Liang, P.~Xu, and B.~Xiang.
\newblock Improve transformer models with better relative position embeddings.
\newblock {\em arXiv preprint arXiv:2009.13658}, 2020.

\bibitem{jia2022detrs}
D.~Jia, Y.~Yuan, H.~He, X.~Wu, H.~Yu, W.~Lin, L.~Sun, C.~Zhang, and H.~Hu.
\newblock Detrs with hybrid matching.
\newblock {\em arXiv preprint arXiv:2207.13080}, 2022.

\bibitem{li2022dn}
F.~Li, H.~Zhang, S.~Liu, J.~Guo, L.~M. Ni, and L.~Zhang.
\newblock Dn-detr: Accelerate detr training by introducing query denoising.
\newblock {\em arXiv preprint arXiv:2203.01305}, 2022.

\bibitem{li2022exploring}
Y.~Li, H.~Mao, R.~Girshick, and K.~He.
\newblock Exploring plain vision transformer backbones for object detection.
\newblock {\em arXiv preprint arXiv:2203.16527}, 2022.

\bibitem{lin2017feature}
T.-Y. Lin, P.~Doll{\'a}r, R.~Girshick, K.~He, B.~Hariharan, and S.~Belongie.
\newblock Feature pyramid networks for object detection.
\newblock In {\em Proceedings of the IEEE conference on computer vision and
  pattern recognition}, pages 2117--2125, 2017.

\bibitem{lin2017focal}
T.-Y. Lin, P.~Goyal, R.~Girshick, K.~He, and P.~Doll{\'a}r.
\newblock Focal loss for dense object detection.
\newblock In {\em Proceedings of the IEEE international conference on computer
  vision}, pages 2980--2988, 2017.

\bibitem{liu2022dab}
S.~Liu, F.~Li, H.~Zhang, X.~Yang, X.~Qi, H.~Su, J.~Zhu, and L.~Zhang.
\newblock Dab-detr: Dynamic anchor boxes are better queries for detr.
\newblock {\em arXiv preprint arXiv:2201.12329}, 2022.

\bibitem{liu2016ssd}
W.~Liu, D.~Anguelov, D.~Erhan, C.~Szegedy, S.~Reed, C.-Y. Fu, and A.~C. Berg.
\newblock Ssd: Single shot multibox detector.
\newblock In {\em ECCV}, pages 21--37. Springer, 2016.

\bibitem{liu2022swin}
Z.~Liu, H.~Hu, Y.~Lin, Z.~Yao, Z.~Xie, Y.~Wei, J.~Ning, Y.~Cao, Z.~Zhang,
  L.~Dong, et~al.
\newblock Swin transformer v2: Scaling up capacity and resolution.
\newblock In {\em CVPR}, pages 12009--12019, 2022.

\bibitem{liu2021swin}
Z.~Liu, Y.~Lin, Y.~Cao, H.~Hu, Y.~Wei, Z.~Zhang, S.~Lin, and B.~Guo.
\newblock Swin transformer: Hierarchical vision transformer using shifted
  windows.
\newblock In {\em ICCV}, pages 10012--10022, 2021.

\bibitem{meng2021CondDETR}
D.~Meng, X.~Chen, Z.~Fan, G.~Zeng, H.~Li, Y.~Yuan, L.~Sun, and J.~Wang.
\newblock Conditional detr for fast training convergence.
\newblock In {\em Proceedings of the IEEE International Conference on Computer
  Vision (ICCV)}, 2021.

\bibitem{ouyangzhang2022nms}
J.~Ouyang-Zhang, J.~H. Cho, X.~Zhou, and P.~Kr{\"a}henb{\"u}hl.
\newblock Nms strikes back.
\newblock {\em arXiv preprint arXiv:2212.06137}, 2022.

\bibitem{radford2018improving}
A.~Radford, K.~Narasimhan, T.~Salimans, I.~Sutskever, et~al.
\newblock Improving language understanding by generative pre-training.
\newblock 2018.

\bibitem{Redmon_2016_CVPR}
J.~Redmon, S.~Divvala, R.~Girshick, and A.~Farhadi.
\newblock You only look once: Unified, real-time object detection.
\newblock In {\em Proceedings of the IEEE Conference on Computer Vision and
  Pattern Recognition (CVPR)}, June 2016.

\bibitem{ren2015faster}
S.~Ren, K.~He, R.~Girshick, and J.~Sun.
\newblock Faster r-cnn: Towards real-time object detection with region proposal
  networks.
\newblock {\em Advances in neural information processing systems}, 28, 2015.

\bibitem{shao2019objects365}
S.~Shao, Z.~Li, T.~Zhang, C.~Peng, G.~Yu, X.~Zhang, J.~Li, and J.~Sun.
\newblock Objects365: A large-scale, high-quality dataset for object detection.
\newblock In {\em Proceedings of the IEEE/CVF international conference on
  computer vision}, pages 8430--8439, 2019.

\bibitem{shaw2018self}
P.~Shaw, J.~Uszkoreit, and A.~Vaswani.
\newblock Self-attention with relative position representations.
\newblock {\em arXiv preprint arXiv:1803.02155}, 2018.

\bibitem{tan2020efficientdet}
M.~Tan, R.~Pang, and Q.~V. Le.
\newblock Efficientdet: Scalable and efficient object detection.
\newblock In {\em Proceedings of the IEEE/CVF conference on computer vision and
  pattern recognition}, pages 10781--10790, 2020.

\bibitem{teed2020raft}
Z.~Teed and J.~Deng.
\newblock Raft: Recurrent all-pairs field transforms for optical flow.
\newblock In {\em Computer Vision--ECCV 2020: 16th European Conference,
  Glasgow, UK, August 23--28, 2020, Proceedings, Part II 16}, pages 402--419.
  Springer, 2020.

\bibitem{touvron2021training}
H.~Touvron, M.~Cord, M.~Douze, F.~Massa, A.~Sablayrolles, and H.~J{\'e}gou.
\newblock Training data-efficient image transformers \& distillation through
  attention.
\newblock In {\em International conference on machine learning}, pages
  10347--10357. PMLR, 2021.

\bibitem{vaswani2017attention}
A.~Vaswani, N.~Shazeer, N.~Parmar, J.~Uszkoreit, L.~Jones, A.~N. Gomez,
  {\L}.~Kaiser, and I.~Polosukhin.
\newblock Attention is all you need.
\newblock {\em Advances in neural information processing systems}, 30, 2017.

\bibitem{wang2022internimage}
W.~Wang, J.~Dai, Z.~Chen, Z.~Huang, Z.~Li, X.~Zhu, X.~Hu, T.~Lu, L.~Lu, H.~Li,
  et~al.
\newblock Internimage: Exploring large-scale vision foundation models with
  deformable convolutions.
\newblock {\em arXiv preprint arXiv:2211.05778}, 2022.

\bibitem{wang2021anchor}
Y.~Wang, X.~Zhang, T.~Yang, and J.~Sun.
\newblock Anchor detr: Query design for transformer-based detector, 2021.

\bibitem{wu2021rethinking}
K.~Wu, H.~Peng, M.~Chen, J.~Fu, and H.~Chao.
\newblock Rethinking and improving relative position encoding for vision
  transformer.
\newblock In {\em Proceedings of the IEEE/CVF International Conference on
  Computer Vision}, pages 10033--10041, 2021.

\bibitem{xie2022revealing}
Z.~Xie, Z.~Geng, J.~Hu, Z.~Zhang, H.~Hu, and Y.~Cao.
\newblock Revealing the dark secrets of masked image modeling.
\newblock {\em arXiv preprint arXiv:2205.13543}, 2022.

\bibitem{xie2021simmim}
Z.~Xie, Z.~Zhang, Y.~Cao, Y.~Lin, J.~Bao, Z.~Yao, Q.~Dai, and H.~Hu.
\newblock Simmim: A simple framework for masked image modeling.
\newblock {\em arXiv preprint arXiv:2111.09886}, 2021.

\bibitem{xie2022simmim}
Z.~Xie, Z.~Zhang, Y.~Cao, Y.~Lin, J.~Bao, Z.~Yao, Q.~Dai, and H.~Hu.
\newblock Simmim: A simple framework for masked image modeling.
\newblock In {\em CVPR}, pages 9653--9663, 2022.

\bibitem{zhang2022semantic}
G.~Zhang, Z.~Luo, Y.~Yu, J.~Huang, K.~Cui, S.~Lu, and E.~P. Xing.
\newblock Semantic-aligned matching for enhanced detr convergence and
  multi-scale feature fusion.
\newblock {\em arXiv preprint arXiv:2207.14172}, 2022.

\bibitem{zhang2022towards}
G.~Zhang, Z.~Luo, Y.~Yu, Z.~Tian, J.~Zhang, and S.~Lu.
\newblock Towards efficient use of multi-scale features in transformer-based
  object detectors.
\newblock {\em arXiv preprint arXiv:2208.11356}, 2022.

\bibitem{zhang2022dino}
H.~Zhang, F.~Li, S.~Liu, L.~Zhang, H.~Su, J.~Zhu, L.~M. Ni, and H.-Y. Shum.
\newblock Dino: Detr with improved denoising anchor boxes for end-to-end object
  detection.
\newblock {\em arXiv preprint arXiv:2203.03605}, 2022.

\bibitem{zhu2020deformable}
X.~Zhu, W.~Su, L.~Lu, B.~Li, X.~Wang, and J.~Dai.
\newblock Deformable detr: Deformable transformers for end-to-end object
  detection.
\newblock {\em arXiv preprint arXiv:2010.04159}, 2020.

\end{thebibliography}
}

\vspace{2mm}
\section{Supplementary}

\vspace{1mm}
\section*{A. More Plain ViT Results}
Table~\ref{tab:comp_with_vit} reports more comparison results based on the plain ViT. We use the default setup, described in Section 5.4 of the main text, to adopt a MAE~\cite{he2022masked} pre-trained ViT-Base as the backbone and train the model for {$\sim$}$50$ epochs.
According to the results, we observe that (i) our method boosts the plain DETR baseline from $46.5$ AP to $53.8$ AP when only using a global cross-attention scheme to process single-scale feature maps; (ii) our approach outperforms the strong DETR-based object detector, e.g., Deformable DETR \cite{zhu2020deformable}, which uses a local cross-attention scheme to exploit the benefits of multi-scale feature maps.

\begin{table}[t]
\renewcommand{\arraystretch}{1.5}
\centering
{\begin{center}
\tablestyle{4pt}{1.25}
\begin{tabular}{l|cccccc}
    method & AP & AP$_{50}$ & AP$_{75}$ & AP$_{S}$ & AP$_{M}$ & AP$_{L}$ \\
    \shline
    Plain DETR & $46.5$ & $70.2$ & $50.0$ & $26.3$ & $50.2$ & $65.7$ \\
    Deformable DETR\cite{zhu2020deformable} & $52.1$ & $71.6$ & $56.9$ & $33.5$ & $55.2$ & $\bf{69.0}$ \\
    Ours & $\bf{53.8}$ & $\bf{73.4}$ & $\bf{58.9}$ & $\bf{35.9}$ & $\bf{57.0}$ & $68.9$ \\
\end{tabular}
\end{center}}
\caption{{\small{\textbf{Comparison of the plain DETR baseline, Deformable DETR, and the improved plain DETR with a MIM pre-trained ViT-Base backbone.} Our plain DETR with global cross-attention improves the baseline by a huge margin and outperforms the Deformable DETR, which relies on multi-scale features and local cross attention.
}}}
\label{tab:comp_with_vit}
\end{table}

\section*{B. Runtime Comparison with Other Methods}
We further analyze the runtime cost of different cross-attetnion modulations in Table~\ref{tab:runtime}. BoxRPB slightly increases
runtime compared to standard cross-attention, while having
comparable speed to other positional bias methods.

\section*{C. More Details of Local Attention Scheme}
\begin{figure*}[t]
\centering
\begin{subfigure}[b]{0.195\linewidth}
\includegraphics[width=0.99\textwidth]{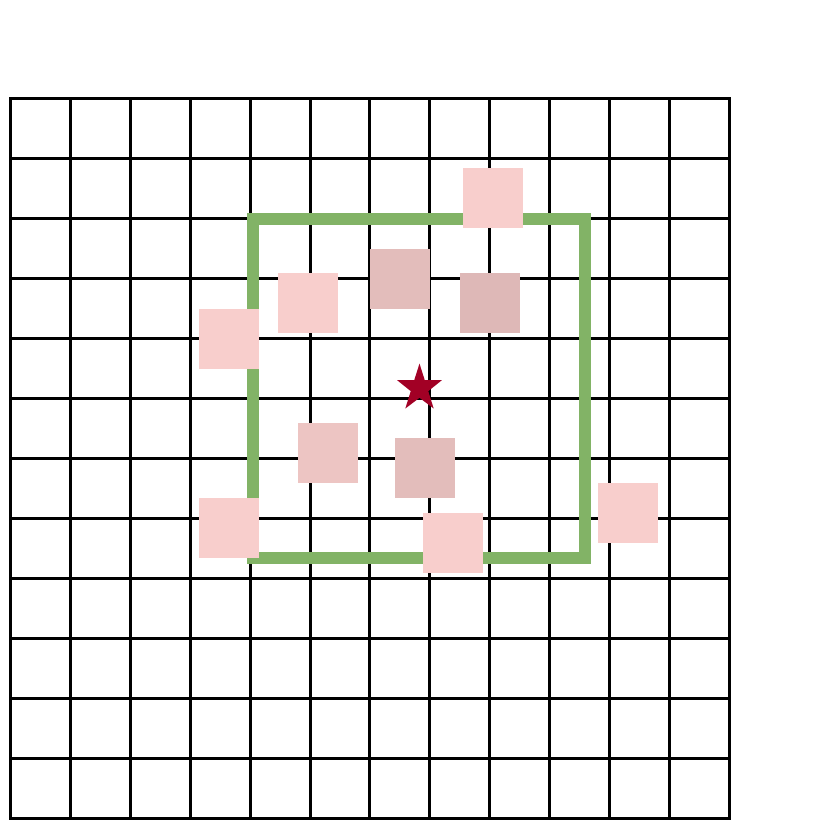}
\caption{\small{Deformable cross-attn.}}
\end{subfigure}
\hfill
\begin{subfigure}[b]{0.195\linewidth}
\includegraphics[width=0.99\textwidth]{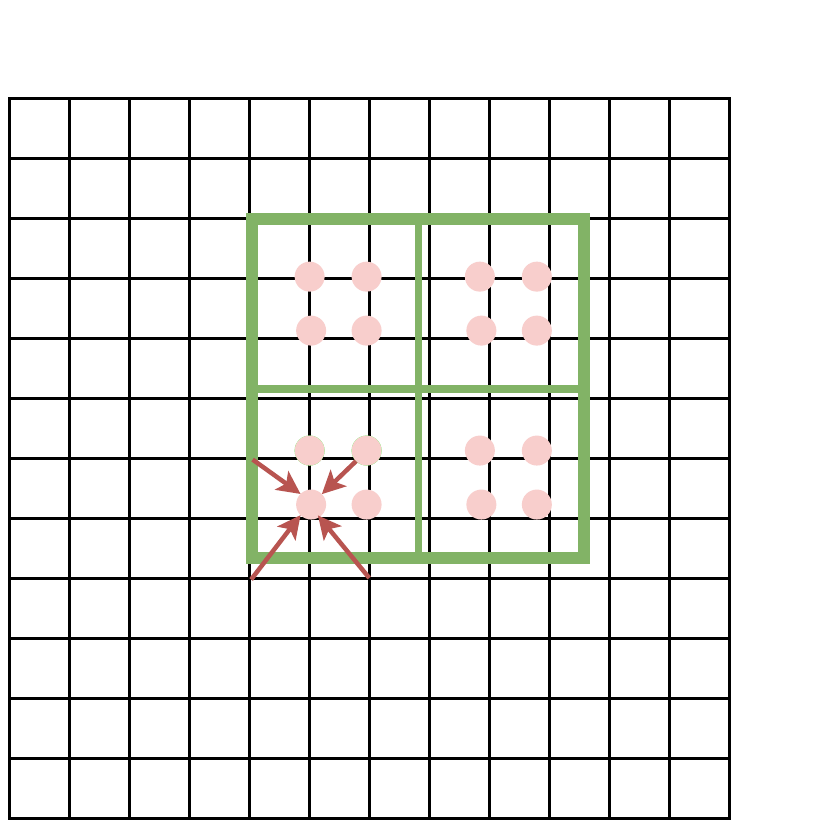}
\caption{\small{RoIAlign}}
\end{subfigure}
\hfill
\begin{subfigure}[b]{0.195\linewidth}
\includegraphics[width=0.99\textwidth]{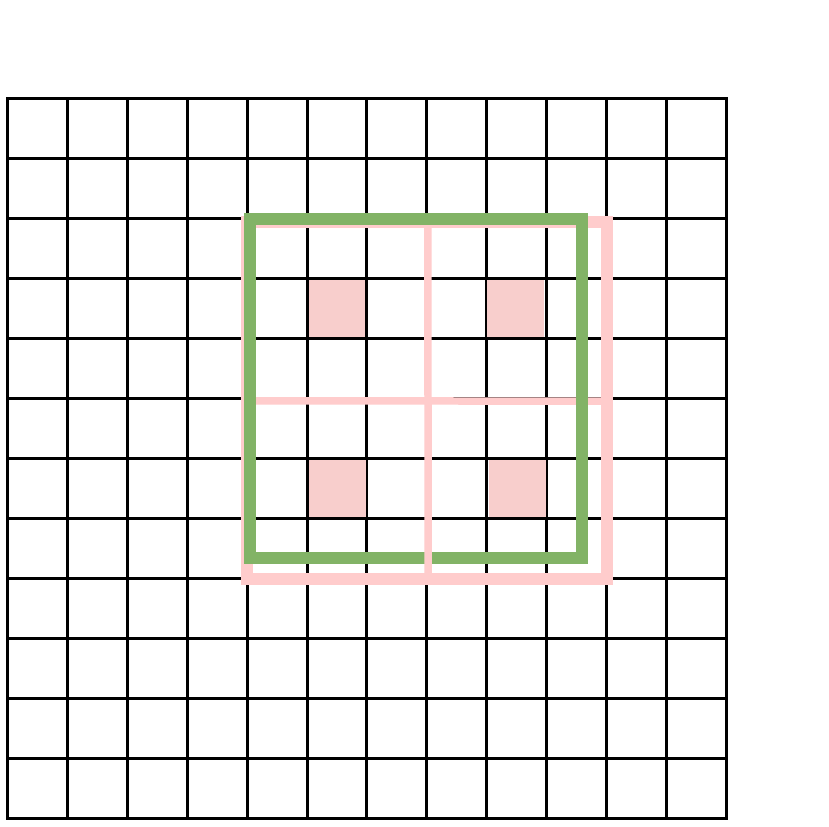}
\caption{\small{RoI Sampling}}
\end{subfigure}
\begin{subfigure}[b]{0.195\linewidth}
\includegraphics[width=0.99\textwidth]{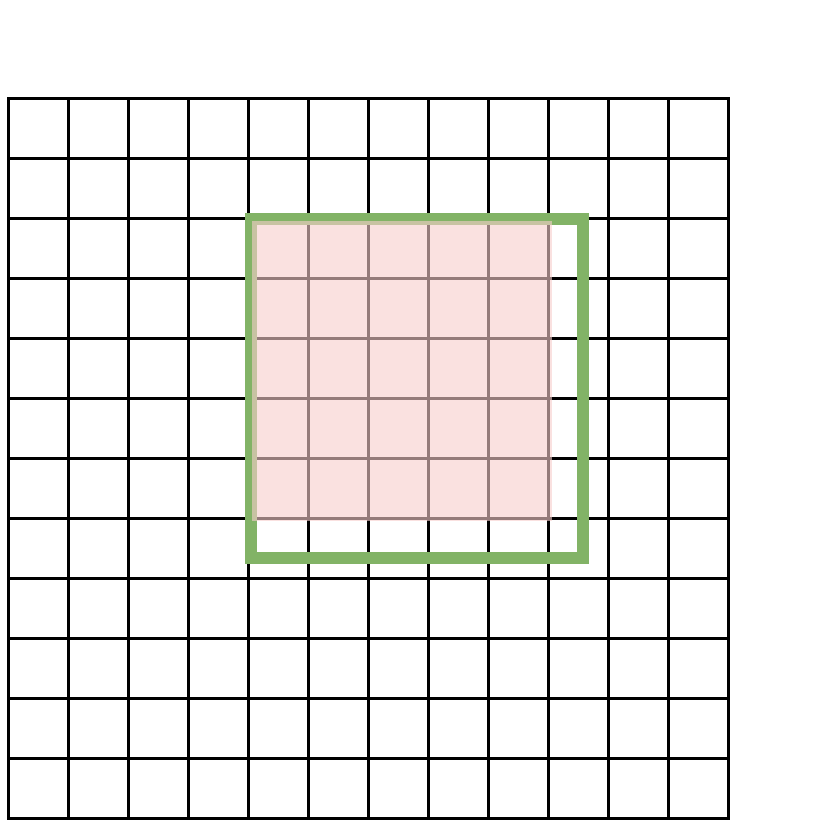}
\caption{\small{Box Mask}}
\end{subfigure}
\begin{subfigure}[b]{0.195\linewidth}
\includegraphics[width=0.99\textwidth]{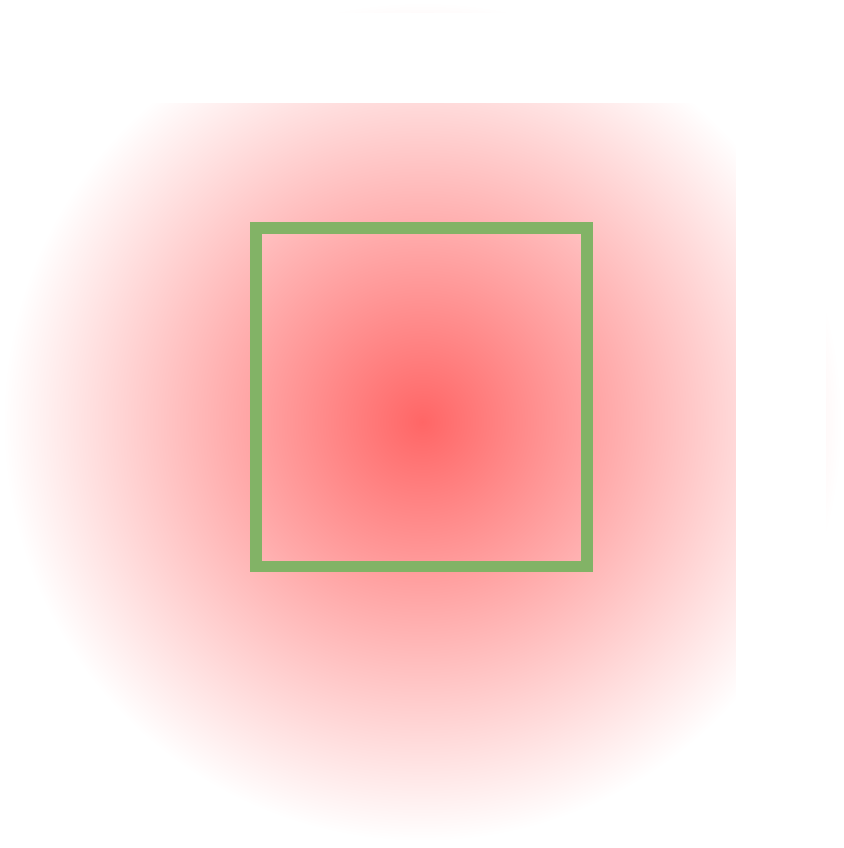}
\caption{\small{Ours}}
\end{subfigure}
\caption{\small{
Illustrating the comparisons between different local cross-attention mechanisms and our global cross-attention schema.
We mark the sampling positions with pink color. The input image is represented by the black grid and the green box is the predicted bounding box from the previous layer. The red star marks the bounding box center. (a) Deformable cross-attention: It learns to sample important positions around the predicted bounding box for the key-value space. (b) RoIAlign: It uses bi-linear interpolation to compute the value of each sampling position in each RoI bin for the key-value space. (c) RoI Sampling: It quantizes the sampling positions to discrete bins and uses them as the key-value space. (d) Box mask: It selects all the positions within the green bounding box as the key-value space. (e) Our method: It improves global cross-attention with BoxRPB, which uses all the positions in the input image as the key-value space. The attention values are indicated by color intensity.
}}
\label{fig:loca_compare}
\end{figure*}

Figure~\ref{fig:loca_compare} shows how our method differs from local cross-attention methods like deformable cross-attention \cite{zhu2020deformable}, RoIAlign \cite{he2017mask}, RoI Sampling (fixed points in the Region of Interest), and box mask from \cite{cheng2021masked}. Most local cross-attention methods need to construct a sparse key-value space with special sampling and interpolation mechanism. Our method uses all image positions as the key-value space and learns a box-to-pixel relative position bias term (gradient pink circular area in (e)) to adjust the attention weights. This makes our method more flexible and general than previous methods.

\begin{table}[t]
\renewcommand{\arraystretch}{1.5}
\centering
{\begin{center}
\tablestyle{4pt}{1.25}
\begin{tabular}{l|cc}
    method &  Training (min/epoch) & Inference (fps) \\
    \shline
    standard cross attn. & $69$ & $9.9$ \\ 
    conditional cross att. & $72$ & $9.5$ \\
    DAB cross attn. & $73$ &  $9.3$  \\
    SMCA cross attn. & $79$ &  $9.6$ \\
    Ours & $75$ & $9.5$ \\
\end{tabular}
\end{center}}
\caption{{\small\textbf{Runtime comparison with local cross-attention scheme.} Global cross-attention with BoxRPB has comparable speed to other positional bias methods.
}}
\label{tab:runtime}
\end{table}

\section*{D. System-level Comparison on COCO val}
Table~\ref{tab:compare_to_sota_coco_val} compares our method with previous state-of-the-art methods when using Swin-Large as the backbone.
With $36$ training epochs, our model achieves $59.8$ AP on COCO \texttt{val}, outperforming DINO-DETR by +$1.3$ AP. With Objects365\cite{shao2019objects365} pre-training, our method gets $63.8$ AP, much higher than DINO-DETR. These results show that, with our approach, the improved plain DETR can achieve competitive performance without intrinsic limitations.

\begin{table*}[t]
\begin{minipage}[t]{1\linewidth}
\newcommand{\blue}[1]{\textcolor{blue}{#1}}
    \definecolor{deepgreen}{rgb}{0.07, 0.53, 0.03}
    \centering
    \setlength{\tabcolsep}{8pt}
    \renewcommand{\arraystretch}{1.5}
    \footnotesize
    \resizebox{0.99\linewidth}{!}
    {
        \begin{tabular}{l|c|c|c|c|cccccc}
            method     & framework & extra data & \#params & \#epoch & AP & AP$_{50}$ & AP$_{75}$ & AP$_{S}$ & AP$_{M}$ & AP$_{L}$ \\
            \shline
            Swin ~\cite{liu2021swin} & HTC & N/A & 284M & $72$ & {$57.1$} & $75.6$ & $62.5$ & $42.4$ & $60.7$ & $71.1$ \\
            Group-DETR~\cite{chen2022group}  & DETR & N/A & $\geq$218M & $36$ & {$58.4$} & $-$ & $-$ & {$41.0$} & {$62.5$} & {$73.9$} \\
            $\mathcal{H}$-Deformable-DETR~\cite{jia2022detrs} & DETR & N/A & 218M & $36$ & {$57.8$} & $76.5$ & $63.7$ & $42.3$ & $61.8$ & $73.1$ \\
            DINO-DETR~\cite{zhang2022dino}  & DETR & N/A & 218M  & $36$ & {$58.5$} & {$77.0$} & {$64.1$} & {$41.5$} & {$62.3$} & {$74.0$} \\
            Ours$^*$ & DETR & N/A & 228M & $36$ & $59.8$ & $78.8$ & $66.0$ & $45.5$ & $63.4$ & $74.2$ \\
            \hline
            DINO-DETR~\cite{zhang2022dino}$^*$  & DETR & O365 & 218M  & $26+18$ & {$63.2$} & {$-$} & {$-$} & {$-$} & {$-$} & {$-$} \\
            Ours$^*$ & DETR & O365 & 228M & $24+24$ & $\bf{63.8}$ & $\bf{81.9}$ & $\bf{70.6}$ & $\bf{50.9}$ & $\bf{67.8}$ & $\bf{77.1}$ \\
        \end{tabular}
    }
    \caption{\small{System-level comparisons with the state-of-the-art methods on COCO \texttt{val}. All methods adopt the Swin-Large backbone. The superscript $*$ marks the results with test time augmentation.}
    \label{tab:compare_to_sota_coco_val}
    }
\end{minipage}
\vspace{5mm}
\end{table*}

\end{document}